\documentclass[journal]{IEEEtran}
\IEEEoverridecommandlockouts

\usepackage{cite}
\usepackage{pifont}
\usepackage{amsmath,amssymb,amsfonts}
\usepackage{booktabs, makecell}
\usepackage{algorithmic}
\usepackage{graphicx}
\usepackage{textcomp}
\usepackage{xcolor}
\usepackage{tabularx}
\usepackage{multicol}
\usepackage{multirow}
\usepackage{caption}
\usepackage{threeparttable}
\usepackage{fontawesome5}
\usepackage{float}
\usepackage{tikz}
\usepackage{verbatim}
\usepackage{pgfplots}
\usepackage{url}
\pgfplotsset{compat=1.18}

\usepackage{subcaption}
\newcommand{\coderepo}{\url{https://github.com/sensorlab/bp-free-tsc}}

\def\BibTeX{{\rm B\kern-.05em{\sc i\kern-.025em b}\kern-.08em
T\kern-.1667em\lower.7ex\hbox{E}\kern-.125emX}}

\usepackage[acronym]{glossaries}
\makeglossaries
\newacronym{ml}{ML}{Machine Learning}
\newacronym{bp}{BP}{Backpropagation}
\newacronym{dfa}{DFA}{Direct Feedback Alignment}
\newacronym{drtp}{DRTP}{Direct Random Target Projection}

\begin{document}

\title{Backpropagation-free Deep Learning for Time-Series Classification
}

\author{\IEEEauthorblockN{%
    Miha~Kap\v{s}\IEEEauthorrefmark{1}, Bla\v{z} Bertalani\v{c}\IEEEauthorrefmark{1}, Carolina Fortuna\IEEEauthorrefmark{1}
\\}\IEEEauthorblockA{%
    \IEEEauthorrefmark{1}Department of Communication Systems, Jo\v{z}ef Stefan Institute, Slovenia\\
}
miha.kaps03@gmail.com, \{blaz.bertalanic, carolina.fortuna\}@ijs.si
\\
}

\maketitle

\begin{abstract}
Deep neural networks for time series classification (TSC) are trained almost exclusively by error backpropagation, which imposes update locking and the weight-transport problem. Backpropagation-free learning rules such as Direct Feedback Alignment (DFA), Direct Random Target Projection (DRTP), and the diffusion-inspired NoProp relax these constraints but have been evaluated almost exclusively on 2D image benchmarks. We adapt these three learners to univariate TSC with 1D convolutional architectures and present the first systematic study of them across 112 datasets from the UCR time series archive, with one hyperparameter-optimized configuration per method. The optimized learners reach mean F1 scores of 0.76 to 0.79, below ROCKET (0.83) and InceptionTime (0.84). We propose an analysis framework that relates the per-dataset gap to InceptionTime to dataset properties and tests convolutional feature learning by replacing the flatten readout with global average pooling, measured as a difference-in-differences (DD) against a backpropagation control on the identical architecture. Intra-class temporal variability is the dataset property most consistently associated with the gap for all three learners (partial Spearman $\rho$ from $0.28$ to $0.40$). Pooling harms DFA and DRTP more than backpropagation (DD $-0.025$ and $-0.055$), consistent with near-random convolutional features, whereas NoProp tolerates pooling as backpropagation does (DD $+0.013$). Relative to backpropagation on the same architectures, DRTP reduces training floating-point operations (FLOPs) by 33\% and trains faster only on series longer than about 700 time steps, DFA yields no computational reduction, and NoProp reduces peak memory by 38\% at the cost of 48\% longer training.
\end{abstract}


\section{Introduction}
\label{sec:introduction}

Time series classification (TSC) is a foundational machine learning task, extensively studied and benchmarked~\cite{middlehurst2024bake}. Historically, state-of-the-art accuracy has been achieved by non-neural ensembles such as HIVE-COTE 2.0~\cite{HIVE-COTE2.0}. However, achieving this performance requires significant computational complexity, often making it impractical since training takes hundreds of hours. Consequently, deep learning models, particularly Fully Convolutional Networks (FCN)~\cite{DeepTSC} and InceptionTime~\cite{Inception}, have become the dominant paradigm. These neural methods are highly competitive in accuracy while training orders of magnitude faster, primarily because they learn powerful, task-specific feature representations through end-to-end training.

Despite their representational power, deep neural networks rely heavily on error backpropagation (BP) to learn these features. Backpropagation enforces strict constraints as it requires storing intermediate activations for the backward pass and mandates sequential layer-by-layer weight updates. This creates two fundamental limitations: update locking, which renders the learning process inherently sequential and highly memory-intensive, and the weight transport problem, which requires physically and biologically implausible symmetric forward-backward weight matrices~\cite{DRTP}.

To bypass these computational overheads, non-neural methods like ROCKET~\cite{ROCKET} achieve high computational efficiency by applying a large number of entirely fixed, random convolutional kernels. However, this presents a fundamental trade-off: fixed random features sacrifice the ability to adapt representations to specific datasets or downstream tasks. To bridge this gap, backpropagation-free (BP-free) neural methods, such as Direct Feedback Alignment (DFA)~\cite{DFA}, Direct Random Target Projection (DRTP)~\cite{DRTP}, and the diffusion-inspired NoProp~\cite{NoProp}, have been proposed. These methods decouple layer-wise dependencies, eliminating the need for sequential backward propagation while retaining the ability to learn. However, these methods have been primarily evaluated on 2D image classification benchmarks, such as MNIST and CIFAR-10~\cite{NoProp, DRTP}. Their effectiveness, stability, and computational dynamics on 1D time-series data remain entirely unexplored.

In this paper, we ask whether any of these learners capture task-specific temporal features on 1D data, or whether they instead rely on fixed random features as ROCKET does. We adapt three BP-free paradigms, DFA, DRTP, and NoProp, to univariate sequences and evaluate them against standard backpropagation across 112 datasets from the UCR archive, each with per-method hyperparameter optimization. To test for feature learning directly, we intervene on the readout, replacing the position-sensitive flatten with global average pooling and measuring a difference-in-differences against a backpropagation control on the identical architecture, which cancels the effects of architecture and dataset difficulty. Our findings separate the three methods into two regimes. DFA and DRTP depend more than backpropagation on the flatten readout, whereas NoProp tolerates pooling as backpropagation does. 

The main contributions of this paper are:
\begin{itemize}
  \item We adapt three backpropagation-free learners (DFA, DRTP, NoProp) to univariate TSC with 1D convolutional architectures and conduct the first systematic study of these learners across 112 UCR datasets with per-method hyperparameter optimization, benchmarked against backpropagation and standard TSC baselines.
  \item We propose a new analysis framework designed to characterize where and why the backpropagation-free methods fall short of end-to-end training, and find that for all three backpropagation-free learners, intra-class temporal variability is the most consistent dataset property associated with the gap to InceptionTime.
  \item We quantify training time, FLOPs, peak memory, energy consumption, and CO2 emissions relative to backpropagation on identical architectures. We show that the computational benefits of backpropagation-free learning are conditional: DRTP reduces training FLOPs but is faster only for sufficiently long sequences, DFA provides little computational reduction in the evaluated configuration, and NoProp substantially reduces peak memory at the cost of longer training time.
\end{itemize}

The paper is structured as follows. Section~\ref{sec:related:work} reviews related work. Section~\ref{sec:problem:statement} formalizes the problem and the studied learners while Section \ref{sec:tsc} elaborates on the adaptation for time series classification. Section \ref{subsec:analysis} elaborates on the proposed analysis framework, Section~\ref{sec:methodology} details the methodology, and Section~\ref{sec:results} presents the results. Section~\ref{sec:conclusions} concludes our findings.

\section{Related work}
\label{sec:related:work}
\subsection{Time Series Classification}
Time series classification (TSC) has been extensively studied and evaluated on the UCR archive \cite{UCR}, the primary benchmark, comprising 128 univariate time series datasets. Current state-of-the-art accuracy on the UCR archive is achieved by HIVE-COTE 2.0 \cite{HIVE-COTE2.0}, a successor of HIVE-COTE \cite{HIVE-COTE}, which is a non-neural ensemble of different classifiers over different time series representations. While achieving an impressive 89\% accuracy across 112 selected datasets, HIVE-COTE's computational complexity makes it impractical for most applications, as training an entire archive takes more than 300 hours \cite{HIVE-COTE2.0}.

Deep learning has been highly successful in TSC, notably with Fully Convolutional Networks (FCNs) and ResNets \cite{DeepTSC}. InceptionTime \cite{Inception}, an ensemble of deep convolutional networks built on Inception modules that apply parallel convolutional filters of varying lengths to capture multi-scale temporal patterns, is competitive with HIVE-COTE yet trains orders of magnitude faster. It was further improved with hand-crafted filters \cite{H-Inception} and by combining Inception modules with FCNs \cite{InceptionFCN}.

Deep models, however, depend on backpropagation, whose constraints (reviewed below and formalized in Section~\ref{sec:problem:statement}) have motivated a range of backpropagation-free alternatives that reduce memory overhead and potentially parallelize layer updates \cite{DFA, DRTP, NoProp, FF, LocalError}.

\subsection{Non-neural methods}
Unlike deep learning models, non-neural methods do not rely on gradient computation or iterative weight updates. By shifting the focus from end-to-end representation learning to algorithmically defined feature spaces, these statistical and geometric models eliminate the need for backpropagation while still achieving strong performance in time series classification.

Among traditional approaches, distance-based algorithms are particularly prominent, notably the Dynamic Time Warping distance \cite{DTW_distance} coupled with the 1-nearest neighbor classifier (1-NN DTW) \cite{DTW}. By aligning sequences optimally, DTW is robust to temporal distortions and remains the standard baseline for new algorithms. Its limitations are twofold: it relies on a fixed distance metric with no learned representation, and its inference cost is high, requiring a distance computation for every test-train pair rather than a single forward pass.

Several approaches focus on automated feature extraction within specific representational domains. Fulcher et~al.\ \cite{TSC_feature_based} introduced a highly comparative framework that automatically extracts thousands of diverse features from time-series data. This massive feature space is subsequently reduced using greedy forward feature selection to identify a small, interpretable subset for linear classification. Other approaches include shapelet-based methods that extract or learn discriminative phase-independent subsequences \cite{Shapelets_Grabocka, Shapelets_Lines, Shapelets_Li}, dictionary-based methods such as BOSS \cite{BOSS} that capture frequency patterns, and tree-based algorithms like Time Series Forest \cite{TSForest}. Rather than relying on handcrafted features, these algorithms computationally discover optimal representations from the data.

More recently, ROCKET \cite{ROCKET} and its successor MiniROCKET \cite{MiniROCKET} have emerged as highly scalable alternatives. ROCKET transforms each series with a large ensemble of randomly initialized, frozen convolutional kernels (typically 10,000), extracting only two scalars per kernel: the maximum value and the proportion of positive values. This over-parameterized projection renders complex temporal dynamics linearly separable, so that a simple Ridge Regression classifier suffices. MiniROCKET replaces the random weights with a small deterministic set, yielding up to an order-of-magnitude speedup without sacrificing accuracy.

This decoupling, however, is also the fundamental drawback of non-neural methods: they rely on fixed projections coincidentally capturing discriminative patterns and lack the capacity for true representation learning.

\subsection{Backpropagation-Free Neural Methods}
The methods reviewed above present a fundamental trade-off: non-neural approaches like ROCKET achieve remarkable efficiency by sacrificing task-specific representation learning, whereas deep learning methods like InceptionTime learn powerful representations but are bottlenecked by error propagation.

Standard gradient-based training carries biological, physical, and computational constraints, principally the weight transport problem \cite{WTProblem}, which requires symmetric forward and backward weight matrices, and update locking \cite{ULock}, which forces sequential, memory-intensive layer updates and impedes parallel or distributed training. We formalize both in Section~\ref{sec:problem:statement}. We review the methods that relax them.

An early efficient alternative is the echo state network (ESN) \cite{ESN}. Sharing the paradigm of ROCKET, ESNs map signals into a high-dimensional state space via a fixed, randomly connected recurrent "reservoir", so that only the final readout layer is trained. Recent variants replace the linear readout with convolutional and max-pooling layers, making them more suitable for TSC \cite{ESN_TSC}.

For deep feedforward architectures, several methods retain learned representations while relaxing gradient-based training. Feedback Alignment (FA) \cite{FA} showed that fixed random feedback weights can replace transposed forward weights, addressing weight transport. Direct Feedback Alignment (DFA) \cite{DFA} projects the output error directly to each hidden layer, eliminating sequential backward propagation. Direct Random Target Projection (DRTP) \cite{DRTP} further removes update locking by substituting target labels for computed errors, enabling fully feedforward training. The Forward-Forward algorithm (FF) \cite{FF} dispenses with backpropagation altogether, training each layer with a local goodness objective over two forward passes on real and corrupted data. Most recently, the diffusion-inspired NoProp \cite{NoProp} has each block independently denoise targets, removing cross-layer gradient flow entirely.

These methods have been validated primarily on image benchmarks such as MNIST and CIFAR-10. Their effectiveness on time series remains unexplored.

\section{Problem statement and Learning Rules}
\label{sec:problem:statement}

A univariate time series $X$ is an ordered sequence $[x_1,x_2, \ldots ,x_L]$, where each $x_l$ for $l \in \{1,2, \ldots ,L\}$ represents a real-valued observation at time step $l$. \cite{Ismail_Fawaz_2019}

A class label $Y$ is a categorical variable associated with a time series $X$, where $Y \in \{y_1, y_2, \ldots, y_C\}$ and $C$ denotes the total number of distinct classes.

A dataset $D = \{(X_1, Y_1), (X_2, Y_2), \ldots, (X_N, Y_N)\}$ consists of $N$ pairs $(X_i, Y_i)$, where $X_i$ is a univariate time series and $Y_i$ is its corresponding class label. \cite{Ismail_Fawaz_2019}

The task of time series classification involves learning a function $f$ that maps an input time series to a probability distribution over class labels,
\begin{equation}
    f: X \mapsto P(Y | X)
\end{equation}
such that for each time series $X_i$, the classifier predicts \\
$P(Y_i = y_c | X_i)$ for each class $c \in \{1, 2, \ldots, C\}$.
\cite{Ismail_Fawaz_2019}

\subsection{Learning with Backpropagation}

\begin{figure*}[!htbp]
    \centering
    \includegraphics[width=0.96\linewidth]{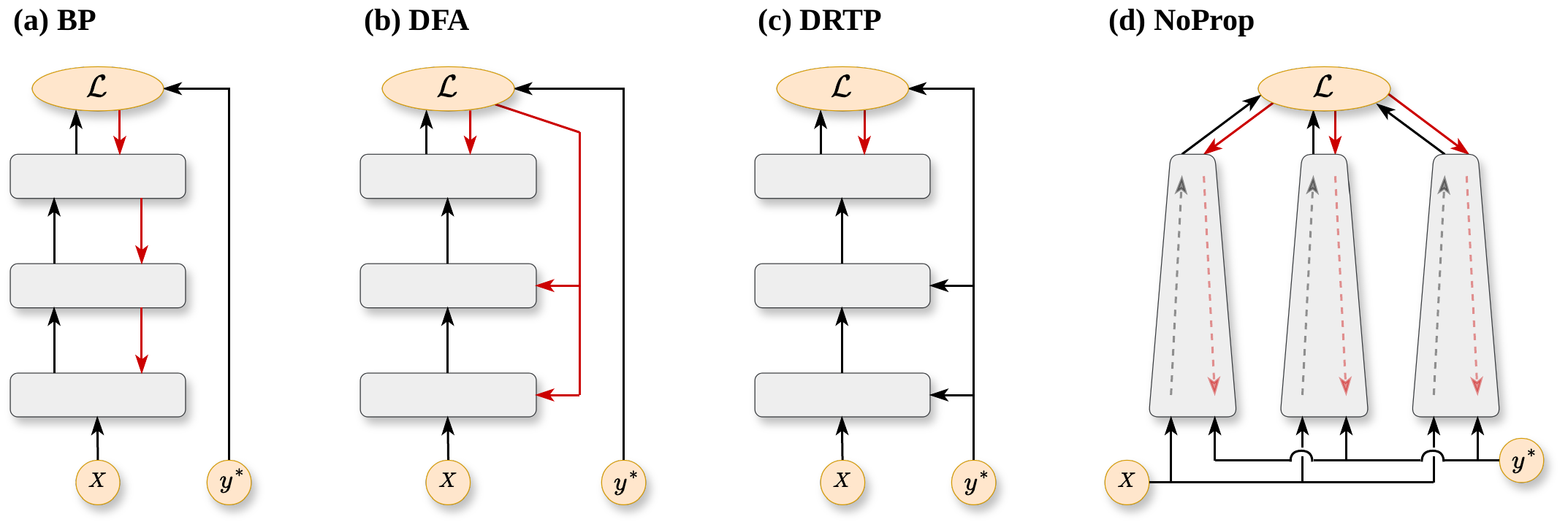}
    \caption{\textbf{Comparison of different error propagation strategies.} Black arrows indicate feedforward pathways and red arrows indicate feedback pathways. Input time series is denoted as $X$, target vector as $y^*$, and loss function as $\mathcal{L}$. \textbf{(a)} Backpropagation: Firstly, the forward pass is performed and error gradients are calculated sequentially for each layer. \textbf{(b)} Direct Feedback Alignment: after forward pass, error is projected independently to each layer. \textbf{(c)} Direct Random Target Projection: target vector is projected to each layer during the forward pass. \textbf{(d)} NoProp: backpropagation occurs in each block independently. There is no gradient flow between the blocks.}
    \label{fig:error_propagation}
\end{figure*}

The classification function $f$ can be implemented with a feedforward neural network, which is most often trained via backpropagation (BP) \cite{BP}, illustrated in Figure~\ref{fig:error_propagation}a. 
Backpropagation computes gradients of the loss function with respect to network parameters by applying the chain rule, propagating error signals backward from the output layer to the input layer.

In a simple feedforward network, a time series $X$ is fed into the input layer and undergoes the forward pass. 
Each layer $k \in \{1,2,\ldots, K\}$ computes the pre-activation $z_{k}$ and activation $a_{k}$ as:
\begin{equation}
    z_{k} = W_{k} a_{k-1} + b_{k}
\end{equation}
\begin{equation}
    a_{k} = \sigma_k(z_{k})
\end{equation}
where $W_{k}$ is the weight matrix, $b_{k}$ is the bias vector, $\sigma_k(\cdot)$ is a non-linear activation function, and $a_{0} = X$.

After the forward pass, the loss is computed. For classification, cross-entropy loss is commonly used:
\begin{equation}
    \mathcal{L}_{CE} = -\sum_{c=1}^{C} y^*_c \log(\hat{y}_c)
\end{equation}
where $y^*$ is the one-hot encoded true label and $\hat{y} = a_{K}$ is the predicted probability distribution over $C$ classes, usually obtained with softmax function.

During the backward pass, the error signal is propagated from the output layer back through the network.
If we choose a softmax activation function in the output layer along with cross-entropy loss, the gradient at the output layer is: 
\begin{equation}
    \delta z_{K} = \frac{\partial\mathcal{L}_{CE}}{\partial z_K} =\hat{y} - y^*
\end{equation}
For hidden layers $k = K-1, K-2, \ldots, 1$, the error signal is computed recursively:
\begin{equation}
    \delta z_{k} = \frac{\partial\mathcal{L}_{CE}}{\partial z_k}= \left( W_{k+1}^T \delta z_{k+1} \right) \odot \sigma_k'(z_{k})
    \label{eq:bp_delta}
\end{equation}
where $z_{k}$ is the pre-activation, $\sigma_k'(\cdot)$ is the derivative of the activation function, and $\odot$ denotes element-wise multiplication.

The weight updates are then computed as:
\begin{equation}
    W_{k} \leftarrow W_k - \lambda \cdot \left(\delta z_{k} a_{k-1}^T \right)
\end{equation}
\begin{equation}
    b_{k} \leftarrow b_k - \lambda \cdot \delta z_{k}
\end{equation}
where $\lambda$ is the learning rate.

\subsection{Backpropagation-Free Learning Methods}
\label{sec:bfa}

To address the physical, biological, and computational constraints of standard error backpropagation, a growing body of research has focused on backpropagation-free learning methods. Figure~\ref{fig:error_propagation} visualizes the structural differences in error routing between standard backpropagation and three prominent alternative strategies. These approaches aim to decouple the layer-wise dependencies that cause update locking and circumvent the weight transport problem by utilizing alternative mechanisms for credit assignment. Broadly, these methods include feedback alignment techniques, which approximate gradient signals with fixed random projections, and localized target propagation or denoising techniques, which train layers independently with local objectives. By relaxing the strict requirements of the exact chain rule, these algorithms enable more biologically plausible architectures and highly parallelizable hardware implementations.

\subsubsection{Direct Feedback Alignment}
Introduced by N{\o}kland et~al., Direct Feedback Alignment (DFA) \cite{DFA} addresses the weight transport
problem by propagating the error signal directly from the output layer to each hidden layer through
fixed random connections, bypassing the layer-by-layer backward pass entirely.

Lillicrap et~al.\ \cite{FA} first demonstrated with Feedback Alignment (FA) that backward weights need not be symmetric to forward weights, replacing $W^T$ from Equation~\ref{eq:bp_delta} with fixed random matrices $B_{k}$:
\begin{equation}
    \delta z_{k} = B_{k} \delta z_{k+1} \odot \sigma_k'(z_{k})
\end{equation}
They have shown that despite random feedback weights, the network learns to adjust its forward weights in such a way that the error signals from FA approximate those of backpropagation, enabling effective learning. 


DFA \cite{DFA} extends this principle by eliminating sequential layer-by-layer error propagation entirely (see Figure~\ref{fig:error_propagation}b). Instead, the output error signal $e = \delta z_K =\hat{y} - y^*$ is projected directly to each hidden layer through independent random matrices:
\begin{equation}
    \delta z_{k} = B_{k} e \odot \sigma_k'(z_{k})
    \label{eq:dfa}
\end{equation}
where $B_{k}$ is a fixed random matrix connecting the output layer directly to layer $k$, and $\sigma_k'(z_{k})$ is the derivative of the activation function. Experimental results \cite{DFA} show that FA and DFA achieve similar performance on the MNIST and CIFAR-10 datasets, slightly lagging behind BP. 

\subsubsection{Direct Random Target Projection (DRTP)}
While DFA eliminates the weight transport, computing the error signal still enforces update locking, as hidden layers must wait for the full forward pass to complete.

Direct Random Target Projection (DRTP), introduced by Frenkel et~al.\ \cite{DRTP}, removes this dependency (as seen in Figure~\ref{fig:error_propagation}c). In classification tasks with one-hot encoded targets, the target vector $y^*$ can serve as a proxy for the error direction. DRTP replaces the output error $e$ with the target label itself:
\begin{equation}
    \delta z_{k} = B_{k} y^* \odot \sigma_k'(z_{k})
    \label{eq:drtp}
\end{equation}
where $B_{k}$ is a fixed random matrix, $y^*$ is the one-hot encoded target vector, and $\sigma_k'(z_{k})$ is the derivative 
of the activation function.

Since $y^*$ is one-hot encoded, the projection $B_{k} y^*$ reduces to simply selecting the column of $B_{k}$ corresponding to the target class. This formulation enables fully feedforward training: each layer computes its weight update using only its local input $a_{k-1}$ and the global target $y^*$, without waiting for downstream computations.

\subsubsection{NoProp}
\label{subsec:noprop}

\begin{figure}[!htbp]
    \centering
    \includegraphics[width=0.96\linewidth]{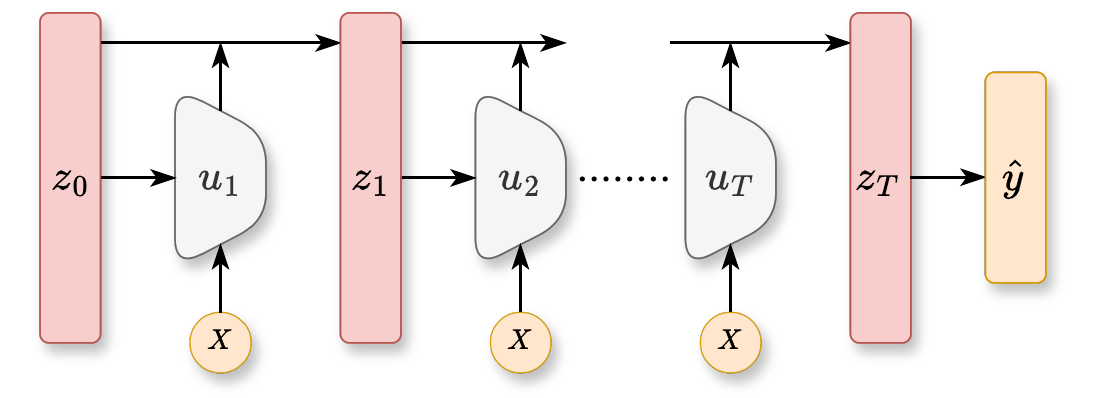}
    \caption{\textbf{Architecture of NoProp network}. The input $z_0$ is sampled from a Gaussian noise distribution and is iteratively transformed into $z_1, \dots, z_T$ via a sequence of learned functions $u_1, \dots, u_T$. Each transformation block incorporates the input time series $X$ as a conditioning signal, guiding the representations through successive refinements until a final class prediction $\hat{y}$ is produced.}
    \label{fig:noprop_architecture}
\end{figure}

Introduced by Li et~al., NoProp \cite{NoProp} takes a different approach to classification. Instead of propagating signals forward and backward across the entire network, NoProp frames learning as a denoising problem inspired by diffusion models and flow matching. Each block independently learns to denoise a noisy version of the target label, using the input data as conditioning. NoProp's architecture is outlined in Figure~\ref{fig:noprop_architecture}. 

During training, the clean embedding of the target label, $u_Y$, is corrupted by a fixed schedule of Gaussian noise to produce noisy representations $z_{t-1}$, sampled as:
\begin{equation}
    z_{t-1} \sim q(z_{t-1} \mid Y) = \mathcal{N}_d(z_{t-1} | 
    \sqrt{\bar{\alpha}_{t-1}} u_Y, 1 - \bar{\alpha}_{t-1}),
\end{equation}
where $u_Y$ is the embedding of class label $Y$ and $\bar{\alpha}_t = \prod_{s=t}^{T} \alpha_s$ controls the noise level.
The noise schedule is designed such that $z_T$ is close to the clean label embedding $u_Y$, while $z_0$ is pure Gaussian noise.

Each block $\hat{u}_{\theta_t}$ is trained independently to predict the clean label embedding, conditioned on the raw time series $X$. The overall NoProp objective combines three terms:
\begin{equation}
\begin{aligned}
    \mathcal{L}_t = 
    & - \log \hat{p}_{\theta_{\text{out}}}(Y \mid z_T) \\[6pt]
    & + D_{\text{KL}}\bigl(q(z_0 \mid Y) \,\|\, p(z_0)\bigr) \\[6pt]
    & + \frac{T}{2} \eta \bigl( \text{SNR}(t) - \text{SNR}(t-1) \bigr) 
      \bigl\| \hat{u}_{\theta_t}(z_{t-1}, X) - u_Y \bigr\|^2
\end{aligned}
\label{eq:noprop_loss}
\end{equation}

The first term is the standard cross-entropy loss, ensuring that the final representation $z_T$ can be correctly classified. The second term regularizes the latent space by encouraging the noise distribution $q(z_0|Y)$ to match the prior $p(z_0) = \mathcal{N}_d(0, I)$. The third term is the core denoising objective: each block learns to predict the clean label embedding $u_Y$ from its noisy input $z_{t-1}$, weighted by the change in signal-to-noise ratio $\text{SNR}(t) = \frac{\bar{\alpha}_t}{1 - \bar{\alpha}_t}$. The hyperparameter $\eta$ controls the relative importance of the denoising objective.

During inference, blocks are chained sequentially. Starting from pure Gaussian noise $z_0$, each block takes the noisy label produced by the previous block, and denoises it according to the diffusion dynamic:
\begin{equation}
    z_t = a_t \hat{u}_{\theta_t} (z_{t-1}, X) + b_t z_{t-1} + \sqrt{c_t}\epsilon_t,
\end{equation}
where $a_t, b_t$ and $c_t$ are parameters defined by the noise schedule. The final output $z_T$ is mapped to class probabilities via a linear layer followed by softmax.

This architecture enables fully parallel training of blocks, as each block can be updated independently without requiring gradient flow across the network.

\section{Proposed Adaptation to Time Series Classification}
\label{sec:tsc}
We adapt and study three backpropagation-free methods discussed in Section~\ref{sec:bfa} on time series classification tasks. All three learning methods avoid the global backward pass, but they differ in how credit is assigned to hidden layers:
\begin{itemize}
  \item \textbf{DFA}~\cite{DFA} propagates the global output error to each layer through fixed random feedback matrices (Eq.~\ref{eq:dfa}).
  \item \textbf{DRTP}~\cite{DRTP} updates each layer toward a fixed random projection of the target (Eq.~\ref{eq:drtp}), using neither the global error nor a backward pass.
  \item \textbf{NoProp}~\cite{NoProp} trains each block independently with standard gradient descent on a local denoising objective (local backpropagation within a block); blocks are chained without any end-to-end backward pass.
\end{itemize}
NoProp therefore performs gradient-based feature learning at the block level, whereas DFA and DRTP replace gradients with random feedback or projections. None of the three performs global backpropagation, so the distinction is the quality of credit assignment. We adapt the three methods to 1D sequences by converting convolutions to 1D operations and matching the input layer to each dataset's length. No other architectural changes are made at the baseline stage, allowing us to assess out-of-the-box performance as a reference point.

\begin{figure}[!htbp]
    \centering
    \includegraphics[width=0.9\linewidth]{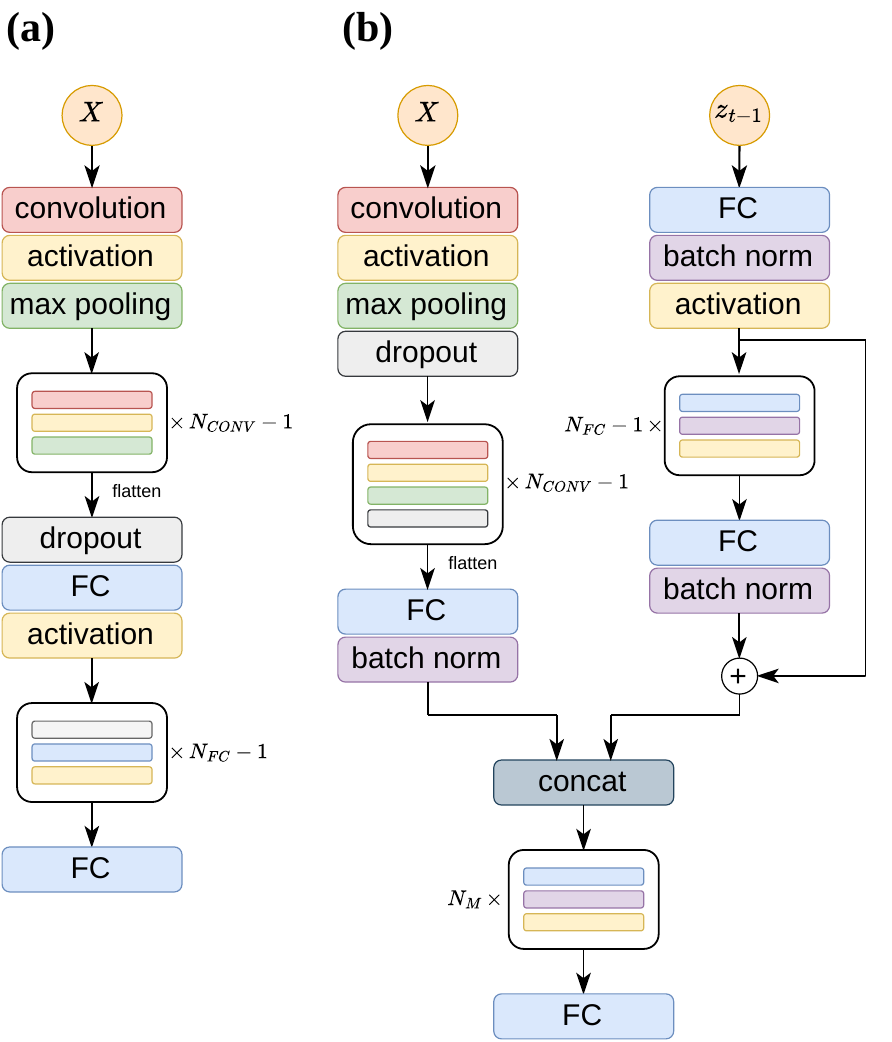}
    \caption{\textbf{Architectures of models used in this study.} \textbf{(a)} Architecture shared by BP, DFA, and DRTP, consisting of convolutional modules followed by fully connected layers. \textbf{(b)} Architecture of a NoProp block, in which the input $X$ is processed through a convolutional stream, while the previous latent state $z_{t-1}$ is processed through a residual fully connected stream. The two streams are concatenated and passed through shared layers.}
    \label{fig:Block_architecture}
\end{figure}

The architectures for all models are illustrated in Figure~\ref{fig:Block_architecture}. For the DFA and DRTP architectures, the network consists of a sequential feature extraction module followed by a classification head. The feature extraction stage utilizes a series of 1D convolutional modules, each containing a convolutional layer, a non-linear activation function, and a max-pooling operation. The outputs are then flattened and passed through a dropout layer. This is followed by a series of fully connected (FC) blocks, each consisting of an FC layer and a non-linear activation. Finally, a FC layer matching the number of target classes is appended to produce the classification output.

The architecture for NoProp consists of a series of independent, locally trained blocks, which inherently removes the requirement for an end-to-end backward pass. Each block processes two distinct inputs. First, the raw time series $X$ passes through a series of convolutional modules, comprising convolution, activation, max-pooling, and dropout. The resulting feature map is flattened, passed through an FC layer, and then batch-normalized. Second, the output representation from the previous block, $z_{t-1}$, is processed through a sequence of fully connected modules consisting of FC layers, batch normalization, and activation functions, followed by an additional FC layer and batch normalization step. Notably, this branch employs a residual connection, adding the output of the initial FC block to the output of the final FC block in this sequence. The processed representations from both branches are then concatenated. Finally, the concatenated features are passed through merging modules (FC, batch normalization, and activation) and a final FC layer to produce the output, from which the $z_t$ for the subsequent block is obtained. Blocks are then chained sequentially as shown in Figure~\ref{fig:noprop_architecture}. 

\section{Proposed Analysis Framework}
\label{subsec:analysis}


To characterize where and why the backpropagation-free methods fall short of end-to-end training, we analyze them along two axes, which dataset properties are associated with their gap to InceptionTime, and whether each method learns task-specific convolutional features.

\subsubsection{Dataset-property characterization}
To identify which dataset conditions are associated with a larger gap to InceptionTime, defined per dataset as $\Delta\textrm{F1} = \textrm{F1}_{\textrm{InceptionTime}} - \textrm{F1}_{\textrm{method}}$, we characterize each dataset by four properties: time series length $L$, number of classes $C$, training-set size $N$, and a within-class temporal-variability metric $D_{L1}$, the mean pairwise L1 distance between standardized training instances of the same class, normalized by length:
\begin{equation}
    D_{L1} = \frac{1}{C}\sum_{c=1}^{C}
    \frac{1}{\binom{|S_c|}{2}}
    \sum_{\substack{X_1,X_2 \in S_c \\ X_1 \neq X_2}}
    \frac{1}{L}\left\| X_1 - X_2 \right\|_1
    \label{eq:l1_distance}
\end{equation}
where $S_c$ is the set of training instances of class $c$. A low $D_{L1}$ indicates that same-class instances are temporally similar and aligned whereas a high value indicates substantial intra-class temporal variability, requiring a model to generalize across temporal shifts and deformations. $D_{L1}$ therefore serves as a proxy for the temporal invariance a dataset demands.

Because $L$, $C$, $N$, and $D_{L1}$ co-vary, we report both zero-order Spearman correlations of each property with $\Delta\textrm{F1}$ and partial Spearman correlations that control for the remaining three. The partial analysis isolates each property's unique contribution and tests whether it retains predictive value once the other properties are held fixed.

\subsubsection{Readout intervention}
To analyze whether each method learns position-invariant temporal features, we intervene on the readout, which maps the convolutional feature maps to the classifier. In their original implementations the three methods adopt a flatten readout, which keeps every time step and is position-sensitive. We replace it with global average pooling (GAP), which averages each channel over time and is position-invariant, such readout appears in end-to-end models such as InceptionTime. The swap removes positional information from the representation the classifier receives, all other components of the optimized configuration are held fixed.

Changing the readout affects classification performance under any learning rule. The difference in F1 between the flatten and GAP readouts therefore reflects both the readout and the learning rule. To separate the two, we train the same optimized architecture with every combination of learning rule $m \in \{\textrm{BP-free}, \textrm{BP}\}$ and readout $r \in \{\textrm{flatten}, \textrm{GAP}\}$, which gives four configurations. The BP-free rule for which the architecture was optimized (DFA, DRTP, or NoProp) is referred to as the native rule, and backpropagation on the same architecture as its control. All four configurations share the architecture and all hyperparameters except the learning rate, which is tuned separately for each configuration on the held-out tuning split. With $\textrm{F1}(m,r)$ the mean F1 on a dataset, the per-rule readout effect is $\Delta_m = \textrm{F1}(m,\textrm{GAP}) - \textrm{F1}(m,\textrm{flatten})$, and the difference-in-differences (DD) is
\begin{equation}
    \textrm{DD} = \Delta_{\textrm{BP-free}} - \Delta_{\textrm{BP}},
    \label{eq:dd}
\end{equation}
computed per dataset. Subtracting the backpropagation effect cancels every factor shared by the four cells (readout main effect, architecture and dataset difficulty) and retains the learning-rule\,$\times$\,readout interaction: $\textrm{DD}<0$ indicates that pooling harms the native rule more than backpropagation. We test $\Delta_{\textrm{BP-free}}$, $\Delta_{\textrm{BP}}$, and DD against zero with the Wilcoxon signed-rank test, Holm-corrected across the three methods.

\subsubsection{Learning dynamics}
To further illustrate how the learning rule shapes feature learning, we track how each layer's weight spectrum evolves over training with \texttt{WeightWatcher}~\cite{Weightwatcher}.
For a given layer, it computes the eigenvalue distribution of the weight matrix and fits a power law with exponent $\alpha$ to its tail. This exponent reflects how strongly the weights have organized into correlated structure. As a layer learns task-specific features, its spectrum develops a heavier tail and $\alpha$ decreases from the random-matrix range toward the heavy-tailed regime. The trajectory of $\alpha$ over training therefore reports the progress of feature learning without reference to labels or held-out data. We record $\alpha$ per epoch for every layer of a single convolutional architecture trained with backpropagation, DFA, and DRTP, on one representative dataset. The convolutional layers are left unfrozen so the trajectory tests whether each rule can train them. Because estimates of $\alpha$ obtained from small 1D convolutional weight matrices are noisy, we interpret only the direction and stability of each trajectory rather than its absolute value.


\section{Methodology}
\label{sec:methodology}
To present the experimental details of our work, we first describe the datasets and preprocessing, followed by the details of model training and results evaluation.
The code and results for all experiments are publicly available.\footnote{\coderepo}
\subsection{Data and Preprocessing}
\label{subsec:data}
This study was evaluated on the University of California, Riverside (UCR) time series classification benchmark archive~\cite{UCR}, which consists of 128 univariate datasets from various domains, including ECG signals, sensor measurements, measurements of motion, signals derived from images, among others.

The initial dataset pool was filtered according to completeness, as some collections contain instances with inconsistent time-series lengths or missing values. To ensure data consistency for both training and testing, the datasets exhibiting these irregularities were removed. This filtering resulted in a final selection of 112 datasets comprising strictly equal-length time series with no missing data. For each dataset, the Appendix
provides the number of training and test instances, the length of the time series, and the number of classes.

\begin{table}[!htbp] 
\setlength{\tabcolsep}{10pt}
\renewcommand{\arraystretch}{1.1}
\centering
\caption{Summary of sample statistics of the 112 UCR archive time series datasets.}
\begin{tabular}{lcccc}
\toprule
\textbf{Statistic} & \textbf{Train} & \textbf{Test} & \textbf{Length} & \textbf{Class} \\
\midrule
Min    & 16      & 20        & 15     & 2     \\
Max    & 8926    & 16\,800   & 2844   & 60    \\
Mean   & 507.02  & 1108.62   & 551.36 & 8.28  \\
Median & 190.5   & 362       & 315    & 3     \\
StdDev & 1175.71 & 2115.24   & 591.69 & 12.47 \\
\bottomrule
\end{tabular}
\label{tab:ucr_stats}
\end{table}

Table \ref{tab:ucr_stats} further presents descriptive statistics across the selected datasets. Each row corresponds to a statistical measure: minimum, maximum, mean, median, and standard deviation (StdDev) and the columns represent dataset attributes: the number of training and test samples, time series length, and number of classes. It can be seen that the training set sizes vary considerably across datasets, spanning from as few as 16 to as many as 8,926 time-series instances. Input lengths also differ widely, ranging from 15 to 2,844 time steps per instance. The number of target classes further adds to this diversity, covering everything from binary classification problems with 2 classes to settings with up to 60 distinct categories. 

In the first step of preprocessing, each time series $X$ in the dataset $D$ was standardized by applying the transformation $x_{t_{std}} = \frac{x_t - \mu_t}{\sigma_t}$ where $\mu_t$ and $\sigma_t$ represent the mean and standard deviation, respectively, calculated for each specific time point $t$ across the train set of series in $D$. Test instances were standardized using these same training-set statistics, so no test information enters the preprocessing. This process centers each time point to a mean of 0 with a standard deviation of 1, providing uniform scaling throughout the dataset. Additionally, each dataset had its class labels adjusted so that they fell in the range $\{0, 1,\dots, C-1\}$, where $C$ is the number of classes in the dataset.

\subsection{Models and Credit Assignment}
\label{sec:architecture}

To contextualize the evaluated backpropagation-free methods, we include two reference classifiers: ROCKET~\cite{ROCKET}, a non-neural method that applies a large bank of fixed random convolutions followed by a trained linear readout, and InceptionTime~\cite{Inception}, a state-of-the-art end-to-end neural model. Both adopt their official, unmodified implementations.

\subsection{Hyperparameter Optimization}
\label{subsec:hpo}

To identify the optimal architecture and hyperparameters for each method adapted in Section \ref{sec:tsc}, we defined a comprehensive search space encompassing both general network parameters and method-specific configurations. Systematic hyperparameter searches were carried out using \texttt{Optuna} \cite{Optuna}, an automated hyperparameter optimization framework. The complete set of evaluated parameters and their respective search bounds are detailed in Table \ref{tab:optuna_parameters}. The search space includes a flag controlling whether the DFA/DRTP convolutional layers are frozen. Whether the optimizer retains or disables convolutional training is therefore itself a learned outcome we report.

\begin{table}[!htbp]
\caption{Hyperparameters and search space for \texttt{Optuna} optimization.}
\centering
\setlength{\tabcolsep}{5pt}
\renewcommand{\arraystretch}{1.1}
\begin{tabular}{lll}
\toprule
\textbf{Parameter} & \textbf{Model} & \textbf{Search space} \\
\midrule
Learning rate       & All      & $\lambda \in [10^{-4}, 10^{-2}]$ \\
Dropout             & All      & $do \in \{0.0, 0.1, 0.2, 0.3, 0.4\}$ \\
\midrule
Conv. modules       & DFA/DRTP & $N_C \in \{1, 2, 3\}$ \\
Base channels       & DFA/DRTP & $ch \in \{32, 64, 128\}$ \\
Kernel size         & DFA/DRTP & $k \in \{5, 9, 17, 33, 65\}$ \\
Freeze conv. layers & DFA/DRTP & $\text{\ding{102}} \in \{\text{True}, \text{False}\}$ \\
FC modules          & DFA/DRTP & $N_{FC} \in \{1, 2\}$ \\
FC base size        & DFA/DRTP & $S_{FC} \in \{256, 512, 1024, 2048\}$ \\
\midrule
NoProp blocks       & NoProp   & $T \in \{1, 2, ... ,12\}$ \\
Conv. modules       & NoProp   & $N_C \in \{1, 2, 3, 4\}$ \\
Base channels       & NoProp   & $ch \in \{16, 32, 64\}$ \\
Kernel size         & NoProp   & $k \in \{5, 9, 13, 21\}$ \\
FC modules          & NoProp   & $N_{FC} \in \{1, 2, 3, 4\}$ \\
Merged FC modules   & NoProp   & $N_M \in \{1, 2, 3\}$ \\
Merged FC base size & NoProp   & $S_M\in \{256, 512\}$ \\
Scaling factor      & NoProp   & $\eta \in [0.1, 1.0]$ \\
Embedding dim       & NoProp   & $emb \in \{0, 64\}$ \\
\bottomrule
\end{tabular}
\label{tab:optuna_parameters}
\end{table}

Hyperparameter tuning was performed on a single stratified split, separate from the splits used for final evaluation. For DFA and DRTP, the automated optimization process was executed for 80 trials, providing sufficient coverage of the search space. For NoProp, the automated search was intentionally restricted to 10 trials, serving purely to identify a reasonable baseline configuration rather than an optimal one, given the method's substantially higher per-trial computational cost. To compensate for this restricted automated search and to achieve thorough hyperparameter coverage for NoProp, we conducted an extensive ablation study. Starting from the Optuna baseline configuration, each hyperparameter was varied independently while holding all others fixed, systematically covering the full search space defined in Table \ref{tab:ablation_parameters}. This one-at-a-time approach provides explicit, interpretable coverage of each parameter dimension. In every case the reported configuration is the one with the highest F1 on the held-out tuning split: the best Optuna trial for DFA and DRTP, and the best ablation setting for NoProp. To examine whether hyperparameter sensitivity varies with temporal complexity, the ablation results are stratified into four groups by time series length: $(0,150]$, $(150,500]$, $(500,1000]$, and $(1000, \infty)$.

\begin{table}[!htbp]
\caption{Hyperparameters utilized for ablation study on NoProp model}
\centering
\renewcommand{\arraystretch}{1.1}
\begin{threeparttable}
\begin{tabular}{lc}
\toprule
\textbf{Hyperparameters} & \textbf{Values} \\
\midrule
Learning rate                         & $\lambda \in \{0.0001, 0.001, 0.01, 0.1\}$          \\
Dropout                               & $do \in \{0.0, 0.1, 0.2, 0.3, 0.4\}$                \\       
Number of NoProp blocks               & $T \in \{1, 3, 5, 10, 15, 20\}$                     \\
Number of convolutional modules       & $N_C \in \{2, 3, 4, 5, 6, 7, 8, 9\}$                \\
Number of filters                     & $ch \in \{8, 16, 32, 64\}$                          \\
Kernel size                           & $k \in \{5, 9, 17, 33, 65, 129\}$                   \\
Number of fully connected layers      & $N_{FC} \in \{2, 3, 4, 5\}$                         \\
Number of fully merged layers         & $N_M \in \{1, 2, 3\}$                               \\
Scaling factor                        & $\eta \in \{0.01, 0.1, 0.7, 1, 10, 100\}$           \\
Embedding dimension                   & $emb \in \{0, 16, 64, 128\}$\tnote{*}               \\
\bottomrule
\end{tabular}
\begin{tablenotes}
\item[*] $emb=0$ represents a scenario where labels are one-hot encoded, embedding dimension equals the number of classes in dataset and is not learnable.
\end{tablenotes}
\end{threeparttable}
\label{tab:ablation_parameters}
\end{table}
\begin{table*}[!thbp]
\caption{Performance of models with architecture as introduced in their respective documentation \cite{DRTP, NoProp}.}
\centering
\setlength{\tabcolsep}{10pt}
\renewcommand{\arraystretch}{1.1}
\begin{tabular}{lcccccccccc}
\toprule
\multirow{2}{*}{\textbf{Model}} & \multicolumn{2}{c}{\textbf{Accuracy}} & \multicolumn{2}{c}{\textbf{Precision}} & \multicolumn{2}{c}{\textbf{Recall}} & \multicolumn{2}{c}{\textbf{F1 score}} & \multicolumn{2}{c}{\textbf{Training time (s)}}\\ 
{} & \textbf{Mean} & \textbf{StdDev} & \textbf{Mean} & \textbf{StdDev} & \textbf{Mean} & \textbf{StdDev} & \textbf{Mean} & \textbf{StdDev} & \textbf{Mean} & \textbf{StdDev}\\
\midrule  
DFA             & 0.748          & 0.231 & 0.731          & 0.249 & 0.722          & 0.248 & 0.719          & 0.252 & 12.19   & 25.50    \\   
DRTP            & 0.744          & 0.217 & 0.725          & 0.236 & 0.715          & 0.236 & 0.712          & 0.239 & \textbf{11.81} & 24.99    \\   
NoProp          & \textbf{0.766} & 0.196 & \textbf{0.769} & 0.181 & \textbf{0.741} & 0.209 & \textbf{0.738} & 0.210 & 423.78 & 908.71   \\   
\bottomrule
\end{tabular}
\label{tab:default_performance}
\end{table*}

\subsection{Training Setup}
\label{subsec:training}

All models were trained for 600 epochs with a batch size of 64 on a single NVIDIA A100 80GB GPU, ensuring a controlled and reproducible computational environment throughout the study. No early stopping was applied, and the weights from the final epoch were used for evaluation.

For NoProp, we adopted the AdamW optimizer with a fixed weight decay of 0.001. The model was trained with a loss function following the formulation described in Equation~\ref{eq:noprop_loss}. A fixed cosine noise schedule was adopted to govern the diffusion process across layers\cite{NoProp, CosineNoise}. The learning rate and the scaling parameter $\eta$ were determined through an ablation study.

For DFA and DRTP, error signal computation follows the formulations described in Equations \ref{eq:dfa} and \ref{eq:drtp}, respectively. Weight updates for both methods were performed using the Nesterov Accelerated Gradient (NAG) optimizer. Cross entropy was used as a loss function. The learning rates for both methods were optimized via the hyperparameter search procedure described above.

\subsection{Evaluation Protocol}
\label{subsec:evaluation}
The UCR archive provides predefined train and test splits for each dataset. To ensure robustness and reduce variance due to data partitioning, each model was trained on 5 additional manually created stratified random splits, maintaining the same train-test ratio as the official split. All results are reported as averages across these 6 splits over 112 datasets from the UCR archive. To ensure a rigorous comparison, the ROCKET baseline was trained and evaluated on the exact same six data splits, utilizing the identical hardware environment to accurately track both predictive performance and computational efficiency metrics. For InceptionTime, we use the results published in its official repository and do not retrain the model on our splits. Because these results were obtained on the official train-test split alone, they are not strictly comparable with those obtained under our six-split protocol, and InceptionTime is therefore treated 
as a reference for state-of-the-art performance.


For evaluation of our models, metrics like accuracy, precision, recall, and F1 score were used, defined as $\textrm{Accuracy}=\frac{\textrm{TP} + \textrm{TN}}{\textrm{TP} + \textrm{TN} + \textrm{FP} + \textrm{FN}}$, $\textrm{Precision} = \frac{\textrm{TP}}{\textrm{TP} + \textrm{FP}}$, $\textrm{Recall} = \frac{\textrm{TP}}{\textrm{TP} + \textrm{FN}}$, and $\textrm{F1 score} = \frac{2 \times \textrm{Precision} \times \textrm{Recall}}{\textrm{Precision} + \textrm{Recall}}$, where TP, TN, FP, and FN denote true positives, true negatives, false positives, and false negatives, respectively. On multiclass datasets, precision, recall, and F1 are macro-averaged, computed per class and averaged with equal weight.

To assess the statistical significance of performance differences across models, we compare model rankings with a Friedman test and pairwise Wilcoxon signed-rank tests with Holm correction, summarized as a critical difference diagram.

Beyond predictive performance, models are further compared across several dimensions of computational efficiency. The total number of floating-point operations (FLOPs) serves as a hardware-agnostic measure of algorithmic complexity, while wall-clock training times reflect practical runtime costs. Using the \texttt{codecarbon} library, total energy consumption and amount of CO\textsubscript{2} emissions for training were tracked. Together, these metrics provide a broader view of the trade-off between classification accuracy and computational overhead.

\section{Results} 
\label{sec:results}
In this section we evaluate DFA, DRTP, and NoProp, adapted to time series classification as described in Section~\ref{sec:tsc}, using the analysis framework proposed in Section~\ref{subsec:analysis} and the experimental protocol described in Section~\ref{sec:methodology}.

\subsection{Classification Performance Analysis of Reference Architectures}
Table \ref{tab:default_performance} presents classification results for each model in its default configuration, as specified in the respective official code or paper \cite{DRTP, NoProp}. Each row corresponds to a distinct model, with the first column indicating the model type. The remaining columns report four predictive performance metrics. For each metric, the score on every dataset is first averaged over the six splits described in Section~\ref{subsec:evaluation}, and the table reports the mean and standard deviation (StdDev) of these 112 per-dataset averages. The final column reports training time per dataset in seconds, aggregated in the same way.

In their reference configurations, the three methods occupy a narrow predictive range. DFA and DRTP achieve similar predictive performance (accuracy 0.75 and 0.74, F1 0.72 and 0.71 respectively) and train in 9.28~s and 8.58~s per dataset. NoProp attains the highest scores (accuracy 0.77, F1 0.74) but at 431.92~s per dataset, roughly $50\times$ slower than DRTP. The reference architectures therefore already expose the central trade-off: NoProp's 0.02--0.03 F1 advantage costs nearly two orders of magnitude in training time, a gap attributable to its $T$ block-local updates (Section~\ref{subsec:noprop}), which our implementation trains sequentially.

\subsection{Hyperparameter Optimization and Ablation Study}
\label{subsec:optimization}

Table~\ref{tab:optuna_optimal} lists the configurations selected by the procedure described in Section~\ref{subsec:hpo}, using the notation of Table~\ref{tab:optuna_parameters}.

\begin{table}[!htbp]
\caption{\textbf{Optimal hyperparameter configurations} obtained by \texttt{Optuna} and additional ablation study for NoProp.}
\centering
\renewcommand{\arraystretch}{1.1}
\begin{tabular}{lc}
\toprule
\textbf{Model} & \textbf{Configuration} \\
\midrule
DFA        & \makecell{$N_C=3, ch=128, k = 5, \text{\ding{102}} = True,N_{FC} = 1,$\\$ S_{FC}=1024,do=0.4, \lambda = 7.1\times10^{-3}$}\\
\midrule
DRTP       & \makecell{$N_C=3, ch=64, k = 17, \text{\ding{102}} = False, N_{FC} = 1,$\\$ S_{FC}=2048, do=0.0, \lambda = 3.7\times10^{-4}$}\\
\midrule
NoProp     & \makecell{$T=10, N_C=6, ch = 16, k=17, N_{FC}=5,$ \\ $N_M=1,do=0.1,\lambda= 1\times10^{-3} , \eta = 0.7, emb = 0$} \\
\bottomrule
\end{tabular}
\label{tab:optuna_optimal}
\end{table}

\begin{figure*}[!tbp]
    \centering
    \includegraphics{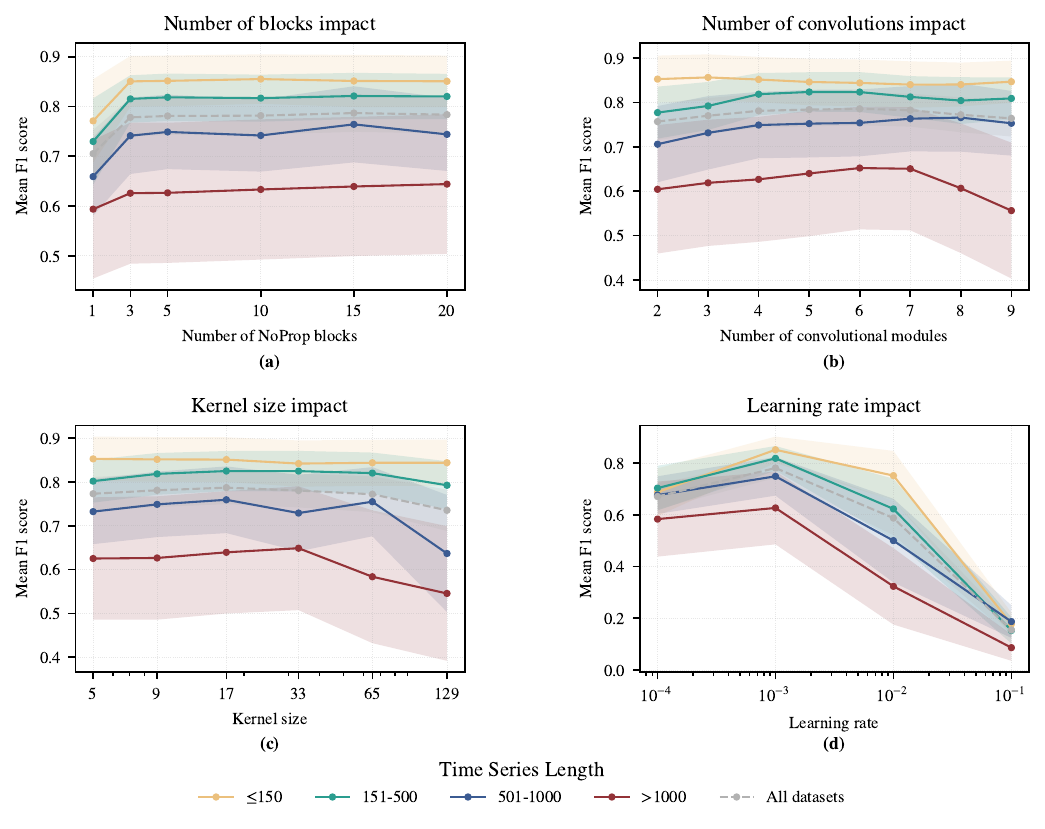}
    \caption{\textbf{Ablation of parameters stratified by time series length.} \textbf{(a)} Adding more than $3$ NoProp blocks results in small performance gains. \textbf{(b)} Performance generally peaks at $N_C=6$. \textbf{(c)} Moderate kernel sizes (between 17 and 33) yield the highest F1 scores on average. \textbf{(d)} Learning rate of $10^{-3}$ provides best performance across all time series lengths. Shaded bands show 95\% confidence intervals of the mean F1 across the datasets in each length group.}
    \label{fig:all_ablation}
\end{figure*} 

DFA and DRTP converged to the same convolutional depth of $N_C = 3$ but differ in their feature extraction and regularization approach. DFA favors wider feature maps with $ch = 128$ channels and a smaller kernel size of $k = 5$, relying on dropout of $do = 0.4$ to regularize the larger parameter space, while DRTP adopts a narrower channel width of $ch = 64$ and a larger kernel size of $k = 17$. Furthermore, DRTP requires no dropout and converges at a lower learning rate of $3.7 \times 10^{-4}$ compared to $7.1 \times 10^{-3}$ for DFA.

Notably, DFA converged to an architecture with randomly initialized, frozen convolutional layers, performing no convolutional feature learning. This matches a known limitation of feedback-alignment methods, the bottleneck effect~\cite{Bottleneck}. Convolutional weight sharing leaves too few free parameters for the forward weights to align with the fixed random feedback, so training these layers yields little benefit over their random initialization and can even reduce accuracy. A trained convolutional layer therefore behaves close to a fixed random projection, which makes the search's choice to freeze it the expected outcome. The same limitation appears for DRTP, which, however, does train its convolutions.

Configuration for NoProp in Table~\ref{tab:optuna_optimal} was selected as the baseline architecture, which we then further optimized by tuning each parameter independently. The effect of some of the varied hyperparameters on classification performance is illustrated in Figure \ref{fig:all_ablation}, with the corresponding search ranges detailed in Table~\ref{tab:ablation_parameters}. Each panel reports the mean F1 score per length group, as defined in Section~\ref{subsec:hpo}, as a function of the varied hyperparameter, alongside the average across all datasets. It can be seen that F1 stratifies by sequence length across all four hyperparameters: shorter datasets score consistently higher than longer ones. The three architectural parameters share a common pattern: their optimum scales with sequence length, and exceeding it degrades the longest sequences most.

\begin{table*}[!tbp]
\caption{\textbf{Comparison of top-performing models.} Along with DFA, DRTP and NoProp, InceptionTime and ROCKET serve as SotA baselines.}
\centering
\setlength{\tabcolsep}{10pt}
\renewcommand{\arraystretch}{1.1}
\begin{tabular}{lcccccccccc}
\toprule
\multirow{2}{*}{\textbf{Model}} & \multicolumn{2}{c}{\textbf{Accuracy}} & \multicolumn{2}{c}{\textbf{Precision}} & \multicolumn{2}{c}{\textbf{Recall}} & \multicolumn{2}{c}{\textbf{F1 score}} & \multicolumn{2}{c}{\textbf{Training time (s)}}\\ 
{} & \textbf{Mean} & \textbf{StdDev} & \textbf{Mean} & \textbf{StdDev} & \textbf{Mean} & \textbf{StdDev} & \textbf{Mean} & \textbf{StdDev} & \textbf{Mean} & \textbf{StdDev}\\
\midrule
DFA             & 0.784 & 0.190 & 0.772 & 0.198 & 0.763 & 0.202 & 0.762 & 0.204 & 14.97  & 27.74   \\ 
DRTP            & 0.804 & 0.183 & 0.792 & 0.194 & 0.783 & 0.198 & 0.782 & 0.199 & 21.23  & 41.28   \\  
NoProp          & 0.810 & 0.178 & 0.804 & 0.178 & 0.788 & 0.195 & 0.787 & 0.197 & 500.01 & 1085.31 \\ 
\midrule
ROCKET          & 0.851 & 0.135 & 0.842 & 0.150 & 0.829 & 0.160 & 0.829 & 0.160 & 3.74    & 16.49   \\
InceptionTime   & 0.851 & 0.148 & 0.844 & 0.160 & 0.835 & 0.168 & 0.839 & 0.164 & 4067.79 & 5139.57 \\

\bottomrule
\end{tabular}
\label{tab:best_performance}
\end{table*}

From Figure~\ref{fig:all_ablation}a it can be seen that for the number of blocks $T$, F1 rises sharply from $T=1$ to $T=3$, then plateaus. Gains beyond $T=5$ are negligible in every group (0.85 for $<$150 timesteps, 0.63--0.65 for $>$1000). As training time scales linearly with $T$, $T=10$ was retained as a conservative upper bound. For convolutional depth $N_C$, Figure~\ref{fig:all_ablation}b shows that the optimal depth increases with length, from $N_C=3$ ($<$150) to $N_C=6$ (150--500) and $N_C=8$ (500--1000). The all-dataset mean peaks at $N_C=6$. Sequences $>$1000 degrade beyond $N_C=7$, consistent with overfitting at high depth. From Figure~\ref{fig:all_ablation}c plotting kernel size $k$ dependency, it can be seen that short sequences are insensitive to $k$ (F1 $\approx$ 0.85 across $k \in \{5,\ldots,129\}$); longer sequences favor wider kernels up to an optimum ($k=17$ for 150--500, $k=33$ for $>$1000) before degrading sharply at $k=129$. Finally, unlike the former parameters, learning rate $\lambda$ dependency in Figure~\ref{fig:all_ablation}d shows a length-independent optimum at $10^{-3}$. Rates of $10^{-2}$ and $10^{-1}$ destabilize block-wise training, while $10^{-4}$ underconverges.

\subsection{Optimized Model comparison}

Table~\ref{tab:best_performance} reports the optimized configurations against ROCKET~\cite{ROCKET} (non-neural baseline) and InceptionTime~\cite{Inception} (neural baseline). Optimization raises every BP-free method by 0.04--0.07 F1 over its default configuration.

\begin{figure}[!htbp]
    \centering
    \includegraphics{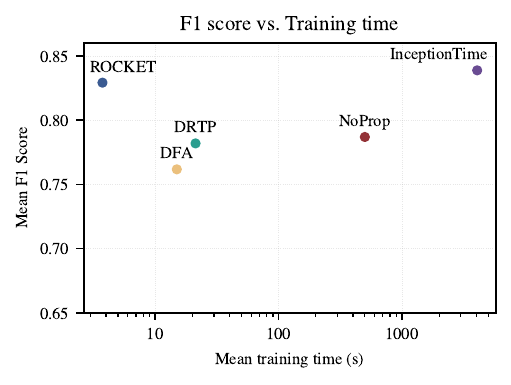}
    \caption{\textbf{Tradeoff between training time and F1 score.} Each point represents a model.}
    \label{fig:Tradeoff}
\end{figure}

It can be seen that among BP-free methods, NoProp leads with F1 around 0.79 and accuracy around 0.81, DRTP achieves slightly lower F1 score of 0.78 and DFA trails at F1 0.76. The methods diverge in compute cost rather than accuracy: DFA and DRTP train in 14.97~s and 21.23~s respectively, whereas NoProp requires 500.01~s, a $23\times$ increase compared to DRTP at comparable F1 score. Both state-of-the-art baselines exceed all BP-free methods on every predictive metric and lie at opposite ends of the trade-off between accuracy and training time, as shown in Figure~\ref{fig:Tradeoff}. InceptionTime reaches an F1 score of 0.84 in 4067.79~s, whereas ROCKET attains 0.83 in 3.74~s and is therefore both faster and more accurate than every BP-free method. The BP-free methods come within 0.04--0.08 F1 of the baselines. Among them, DRTP offers the best balance between accuracy and training time, with an F1 score within 0.06 of InceptionTime at about $190\times$ lower training cost.

\begin{figure}[!htbp]
    \centering
    \includegraphics[width=\linewidth]{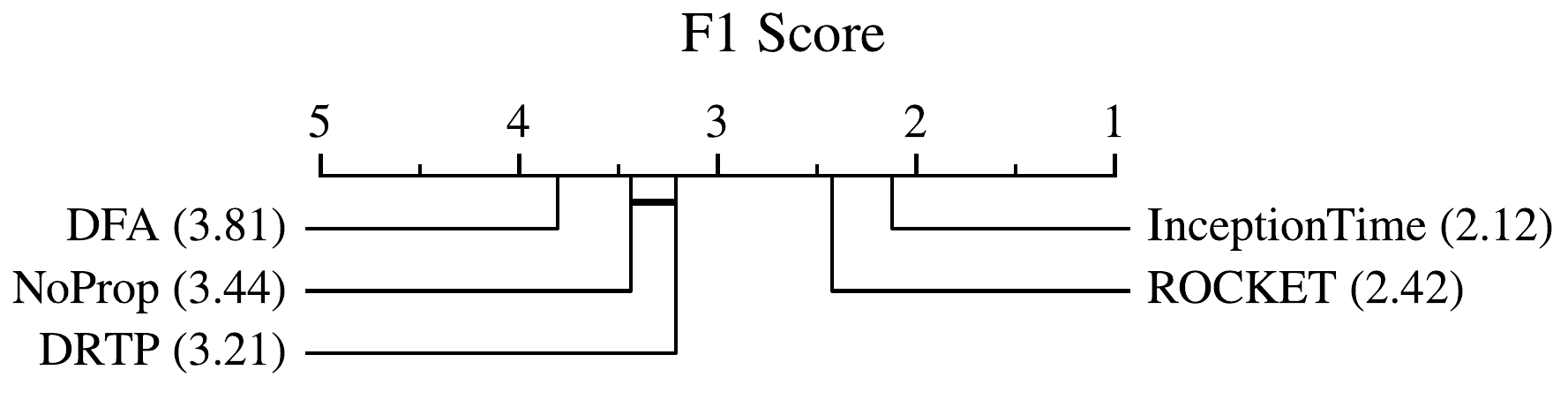}
    \caption{\textbf{Critical difference diagram.} Models whose performance differences are not statistically significant are connected.}
    \label{fig:cdd}
\end{figure}

The critical difference diagram in Figure~\ref{fig:cdd}, obtained with the statistical testing procedure of Section~\ref{subsec:evaluation}, confirms these rankings. DRTP and NoProp do not differ significantly with ranks 3.21 and 3.45. DFA trails with rank 3.81. InceptionTime holds the top rank (2.07), with a significant advantage over ROCKET and all three BP-free methods.


\subsection{Diagnosing the Performance Gap}
\begin{figure*}[!htbp]
    \centering
    \includegraphics{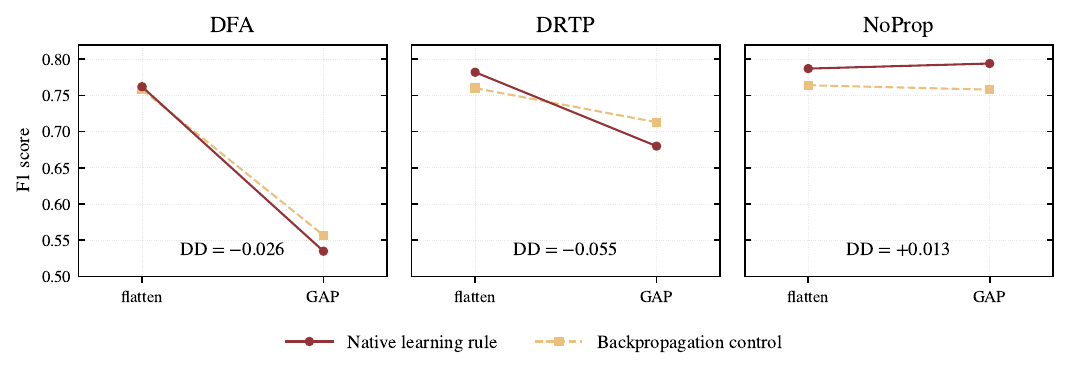}
    \caption{\textbf{Readout intervention as an interaction plot.} Mean F1 as the readout changes from flatten to global average pooling, for the native learning rule and its backpropagation control on the identical architecture. DFA and DRTP yield a negative difference-in-differences, whereas NoProp's native rule is unharmed.}
    \label{fig:pooling_interaction}
\end{figure*}

The optimized backpropagation-free methods trail InceptionTime on most datasets. To test whether this gap reflects a failure to learn task-specific convolutional features, we apply the proposed interventions described in Section~\ref{subsec:analysis}. Results are presented in Table~\ref{tab:dd} and Figure~\ref{fig:pooling_interaction}.

\setlength{\tabcolsep}{5pt}
\begin{table}[!htbp]
\caption{\textbf{Effect of replacing flatten with global average pooling.} Per-dataset paired readout effect $\Delta = \textrm{F1(GAP)} - \textrm{F1(flatten)}$ under each learning rule, and the difference-in-differences $\mathrm{DD}=\Delta_{\mathrm{native}}-\Delta_{\mathrm{BP}}$. ``deg.'' counts degraded datasets. Bold marks effects significant at $p<0.05$ by the Wilcoxon signed-rank test, Holm-corrected across the three methods.}
\label{tab:dd}
\centering
\renewcommand{\arraystretch}{1.1}
\begin{tabular}{l cc cc c}
\toprule
& \multicolumn{2}{c}{\textbf{Native}} & \multicolumn{2}{c}{\textbf{Backprop.}} & \multirow{2}{*}{\textbf{DD}}\\
\cmidrule(lr){2-3}\cmidrule(lr){4-5}
\textbf{Method} & \textbf{mean $\Delta$} & \textbf{deg.} & \textbf{mean $\Delta$} & \textbf{deg.} & \\
\midrule
DFA    & $\mathbf{-0.226}$ & 101/112 & $\mathbf{-0.201}$ & 99/112 & $\mathbf{-0.025}$ \\
DRTP   & $\mathbf{-0.101}$ & 93/112  & $\mathbf{-0.046}$ & 65/112 & $\mathbf{-0.055}$ \\
NoProp & $\mathbf{+0.007}$ & 45/112  & $-0.006$ & 58/112 & $\mathbf{+0.013}$ \\
\bottomrule
\end{tabular}
\end{table}

\begin{figure*}[!htbp]
    \centering
    \includegraphics[width=0.95\linewidth]{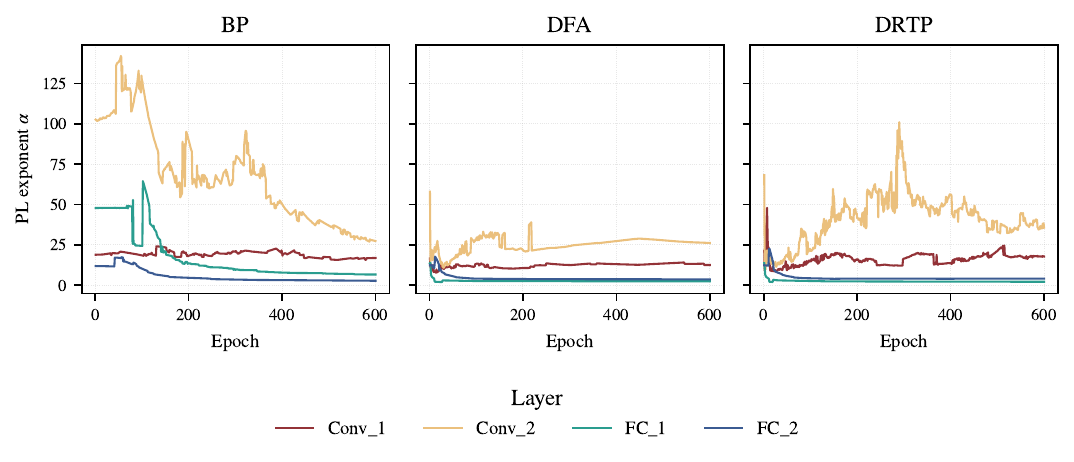}
    \caption{\textbf{Per-layer weight-spectrum exponent $\alpha$ over training on Adiac dataset,} for one convolutional architecture trained with backpropagation, DFA, and DRTP.}
    \label{fig:alpha_grid}
\end{figure*}

Replacing flatten with global average pooling reduces native F1 by $0.226$ for DFA and $0.101$ for DRTP, degrading $101/112$ and $93/112$ datasets, while for NoProp native F1 increases by $0.007$. The difference-in-differences is negative for DFA ($-0.025$) and DRTP ($-0.055$) and positive for NoProp ($+0.013$). The difference-in-differences separates the three methods into two regimes. For DRTP, pooling harms the native rule more than its backpropagation control. DRTP relies on the position-sensitive flatten readout and loses more accuracy than backpropagation when pooling removes it, the behaviour of a random-feature machine rather than a feature learner. DFA shows the same flatten dependence to a greater extent in its native readout effect, yet its difference-in-differences is attenuated because its convolutional layers are frozen. With frozen convolutions, DFA and its backpropagation control both train only the fully connected classification head on identical random features, and global average pooling therefore degrades the control to a similar extent. NoProp forms the second regime. Its native rule is unharmed by pooling, and its backpropagation control on the same architecture is likewise unaffected, so the positive difference-in-differences marks genuine feature learning that tolerates pooling as backpropagation does.

This split follows the bottleneck effect (Section~\ref{subsec:optimization}). Operating on near-random convolutions, DFA and DRTP depend on the high dimensionality of the flatten readout to separate classes, similar to ROCKET, which relies on the high-dimensional output of 10,000 kernels. Global average pooling discards that dimensionality. NoProp, which learns convolutional features, has no such dependence and is unaffected by the readout swap.

Figure~\ref{fig:alpha_grid} reports the per-layer weight-spectrum exponent $\alpha$ over training for backpropagation, DFA, and DRTP on the Adiac dataset, read by trajectory shape as defined in Section~\ref{subsec:analysis}. Under backpropagation the convolutional layers descend toward the heavy-tailed regime, the signature of feature learning, whereas under DFA they stall after an early transient and under DRTP they oscillate without converging. The fully connected layers descend under all three rules. The divergence is confined to the convolutional layers, corroborating that DFA and DRTP do not organize their convolutions into learned features.

\label{subsec:properties}
\begin{figure}[!htbp]
    \centering
    \includegraphics{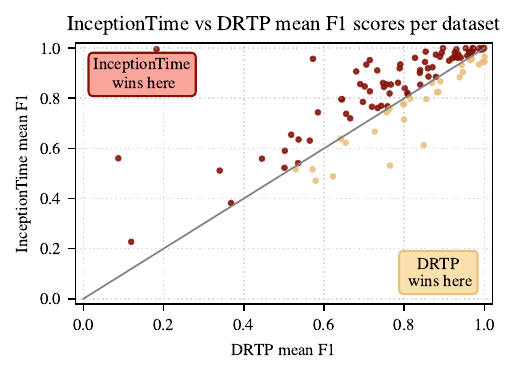}
    \caption{\textbf{Scatter plot comparing mean F1 scores of InceptionTime and DRTP.} Each point corresponds to one dataset.}
    \label{fig:InceptionTime_vs_DRTP}
\end{figure}
\subsection{Dataset-property analysis of the gap}
Figure~\ref{fig:InceptionTime_vs_DRTP} shows the per-dataset F1 scores for DRTP against the state-of-the-art model InceptionTime. Most points lie above the diagonal, indicating that the deficit is present on nearly every dataset but varies widely in magnitude. To identify which dataset properties drive this variation, zero-order and partial Spearman correlations are computed between each dataset property ($L$, $C$, $N$, $D_{L1}$) and each method's per-dataset gap relative to InceptionTime as reported in Table~\ref{tab:partial}.

\setlength{\tabcolsep}{4pt}
\begin{table}[!htbp]
\caption{\textbf{Zero-order and partial Spearman correlations} between dataset properties and the per-dataset gap to InceptionTime. Bold marks Holm-significant partial correlations ($p<0.05$).}
\label{tab:partial}
\centering
\renewcommand{\arraystretch}{1.15}
\begin{tabular}{l cc cc cc}
\toprule
& \multicolumn{2}{c}{\textbf{DFA}} & \multicolumn{2}{c}{\textbf{DRTP}} & \multicolumn{2}{c}{\textbf{NoProp}}\\
\cmidrule(lr){2-3}\cmidrule(lr){4-5}\cmidrule(lr){6-7}
\textbf{Property} & zero & partial & zero & partial & zero & partial\\
\midrule
$L$      & $0.29$  & $\mathbf{0.28}$  & $0.20$  & $0.24$  & $0.19$  & $0.16$\\
$C$      & $0.22$  & $0.02$           & $0.12$  & $-0.15$ & $0.20$  & $0.06$\\
$N$      & $-0.13$ & $0.06$           & $-0.07$ & $0.22$  & $-0.22$ & $-0.03$\\
$D_{L1}$ & $0.36$  & $\mathbf{0.28}$  & $0.36$  & $\mathbf{0.40}$ & $0.43$ & $\mathbf{0.29}$\\
\bottomrule
\end{tabular}
\end{table}

For all three methods, $D_{L1}$ is a Holm-significant partial predictor of the gap with partial Spearman correlations of $\rho=0.28$, $0.40$, and $0.29$ for DFA, DRTP and NoProp respectively. For DRTP and NoProp it is the only one, whereas for DFA, sequence length $L$ additionally survives correction (partial correlation $\rho=0.28$). Neither class count $C$ nor training-set size $N$ is significantly associated with the gap for any method. Intra-class variability is thus the dataset property most associated with the gap, and it holds for all three methods alike. Because NoProp learns convolutional features yet degrades with variability just as DFA and DRTP do, high intra-class variability is a difficulty shared by the whole family, independent of whether a method learns features.

\subsection{Computational cost}
\label{sec:computational_analysis}
We compare the computational cost of each optimized architecture trained under its native backpropagation-free rule and under standard global backpropagation on the identical architecture. Table~\ref{tab:computation} reports mean training time per dataset, total FLOPs, total energy and CO\textsubscript{2} over the 112 datasets, and peak memory.

\setlength{\tabcolsep}{6pt}
\begin{table*}[!htbp]
\caption{Computational cost of backpropagation-free training versus standard backpropagation for each architecture.}
\centering
\setlength{\tabcolsep}{10pt}
\renewcommand{\arraystretch}{1.1}
\begin{tabular}{ccccccc}
\toprule
\makecell{\textbf{Model}\\ \textbf{configuration}} &
\makecell{\textbf{Training}\\ \textbf{method}} &
\makecell{\textbf{Training}\\ \textbf{time} [s]} &
\makecell{\textbf{Total} \\ \textbf{FLOPs}} & 
\makecell{\textbf{Total energy} \\ \textbf{consumed} [kWh]} &
\makecell{\textbf{Total CO\textsubscript{2}} \\ \textbf{emissions} [kg]} &
\makecell{\textbf{Peak memory} \\ \textbf{consumption} [MB]} \\
\midrule
\multirow{2}{*}{DFA}    & Native & 14.97  & $1.07 \times 10^{16}$ & 0.151 & 0.035 & 530.97 \\
                        & BP     & 13.65  & $1.07 \times 10^{16}$ & 0.148 & 0.034 & 521.40 \\
\midrule
\multirow{2}{*}{DRTP}   & Native & 21.23  & $1.56 \times 10^{16}$ & 0.193 & 0.045 & 532.36 \\
                        & BP     & 19.33  & $2.34 \times 10^{16}$ & 0.204 & 0.047 & 574.09 \\
\midrule
\multirow{2}{*}{NoProp} & Native & 500.01 & $4.74 \times 10^{15}$ & 3.963 & 0.854 & 132.79 \\
                        & BP     & 337.80 & $4.74 \times 10^{15}$ & 4.957 & 1.146 & 215.07 \\
\bottomrule

\end{tabular}
\label{tab:computation}
\end{table*}

\begin{figure}[!htbp]
    \centering
    \includegraphics{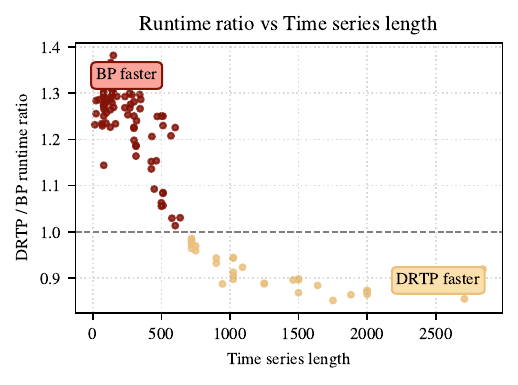}
    \caption{\textbf{DRTP vs. BP runtime ratio as a function of time series length.} Ratios above 1.0 indicate BP is faster; ratios below 1.0 indicate DRTP is faster. The crossover occurs at approximately 700 timesteps.}
    \label{fig:DRTPvBP}
\end{figure}

DRTP is the only method that reduces FLOPs relative to backpropagation. The backward pass holds two matrix products per layer, propagating the error $W_{k+1}^{T}\delta z_{k+1}$ (Equation~\ref{eq:bp_delta}) and forming the weight gradient $\delta z_k a_{k-1}^{T}$, and so costs about twice the forward pass. DRTP replaces the propagated error with a random projection of the target $B_k y^*$ (Equation~\ref{eq:drtp}), whose cost is negligible, because its width is the class count $C$ rather than the larger layer width. Dropping one of the two backward products is a third of the forward-plus-backward total, matching the observed 33\% training FLOPs reduction ($1.56$ versus $2.34\times10^{16}$). The lower FLOP count carries proportional reductions in energy and CO\textsubscript{2}. Peak memory is lower as well (532 versus 574~MB), since DRTP updates during the forward pass and stores no backward graph. Mean training time slightly favours backpropagation (19.33 versus 21.23~s), reflecting \texttt{PyTorch}'s optimized backward pass, but the per-dataset ratio crosses below 1.0 near 700 timesteps (Figure~\ref{fig:DRTPvBP}): on longer sequences the backward pass dominates and DRTP trains faster.

\begin{figure}[!htbp] 
    \centering
    \includegraphics{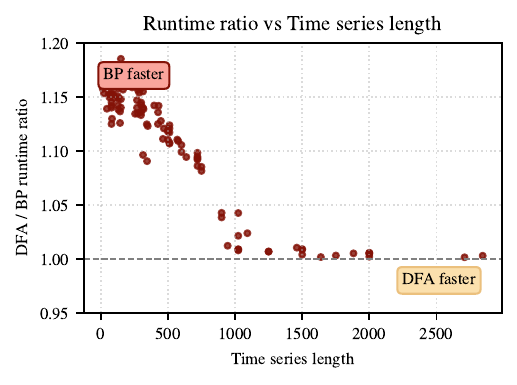}
    \caption{\textbf{DFA vs. BP runtime ratio as a function of time series length.} DFA does not outrun BP for any dataset length.}
    \label{fig:DFAvBP}
\end{figure}

DFA shares this advantage in theory (Equation~\ref{eq:dfa}) but does not realize it, because its optimized architecture freezes the convolutional layers. A frozen convolutional layer is not updated, and with no trainable layer below it, receives no propagated error, so both backward products vanish under both rules. The only trainable part is the fully connected head, where the propagated error already has the output width $C$, so the random projection provides no computational saving. FLOPs, energy, CO\textsubscript{2}, and peak memory are therefore near-identical for DFA and backpropagation ($1.07\times10^{16}$ FLOPs for both). The ratio of DFA to backpropagation training time exceeds 1.0 for all sequence lengths (Figure~\ref{fig:DFAvBP}), indicating that DFA offers no computational advantage over backpropagation in the evaluated configuration.

\begin{figure}[!htbp]
    \centering
    \includegraphics{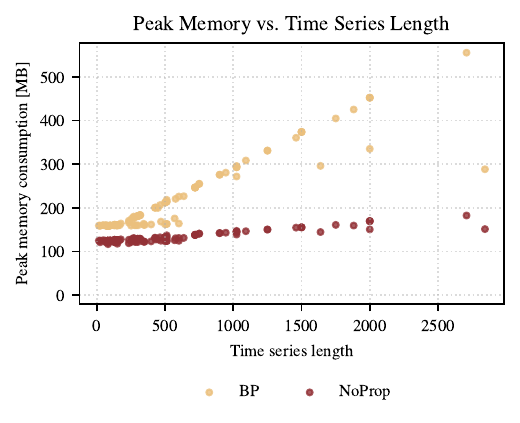}
    \caption{\textbf{NoProp peak memory scaling.} NoProp requires less memory due to its block structure.}
    \label{fig:MemoryxTSL}
\end{figure}

NoProp has equal FLOPs under both rules, as its $T$ block-local updates sum to the cost of one global backward pass. Despite equal FLOPs, backpropagation consumes more energy (4.957 versus 3.963~kWh), emits more CO\textsubscript{2} (1.146 versus 0.854~kg), and requires more peak memory (215.07 versus 132.79~MB), because the global backward pass stores the graph across all blocks. Both scale linearly with sequence length (Figure~\ref{fig:MemoryxTSL}), but NoProp's block-local gradients give a smaller memory coefficient, so its memory advantage grows with length (datasets below the batch size of 64 appear as low-memory outliers). This memory efficiency trades against speed: NoProp trains in 500.01~s under its own rule versus 337.80~s under backpropagation.

\section{Conclusions}
\label{sec:conclusions}
In this study, we explored whether backpropagation-free learners train task-specific temporal features on univariate time series, evaluating DFA, DRTP, and NoProp across 112 UCR datasets against standard baselines and a backpropagation control. The optimized rules reach mean F1 scores of 0.76 to 0.79, below ROCKET (0.83) and InceptionTime (0.84). To separate feature learning from architecture and dataset difficulty, we intervened on the readout, replacing the position-sensitive flatten with global average pooling, and measured a difference-in-differences against a backpropagation control on the identical architecture. The diagnostic split the three methods into two regimes. DFA and DRTP depend more than backpropagation on the position-sensitive flatten readout (DD of $-0.025$ and $-0.055$), consistent with random-feature-like behaviour. DFA's convolutions are frozen and DRTP's show no spectral organization on the inspected dataset, so they lean on the high-dimensional flatten readout and lose accuracy when pooling removes it. NoProp, which trains its convolutions by local gradients, tolerates pooling as backpropagation does (DD of $+0.013$). The deficit to InceptionTime is associated with intra-class temporal variability for all three methods (partial Spearman $\rho$ from $0.28$ to $0.40$), regardless of their regime. These regimes trace a feature-learning versus training-time frontier. DFA and DRTP train in 15 and 21~s per dataset but show no evidence of convolutional feature learning, whereas NoProp requires 500~s, although with fewer FLOPs and lower peak memory than both. Relative to backpropagation on the same architectures, DRTP reduces training FLOPs by 33\% and NoProp reduces peak memory by 38\%, so the computational benefits of backpropagation-free training are conditional on the rule, the sequence length, and the measured resource, and no studied rule matches ROCKET in accuracy or training time. The pooling diagnostic is not specific to these methods and is proposed as a general test of whether a learning rule trains convolutional features, which remains to be validated on other learners.

\section*{Data Availability}
All 112 datasets analysed in this study belong to the publicly available UCR time series classification archive~\cite{UCR}, accessible at \url{https://www.cs.ucr.edu/~eamonn/time_series_data_2018/}. The five additional stratified splits are regenerated from fixed random seeds by the released code, available at \coderepo.

\section*{Acknowledgment}
This work was supported by the Slovenian Research Agency (P2-0016).

\ifCLASSOPTIONcaptionsoff
  \newpage
\fi
\bibliographystyle{IEEEtran}
\bibliography{IEEEabrv,IEEEexample}

\onecolumn
\section*{Appendix}
\label{sec:appendix}
{\small
\captionof{table}{Summary of average F1-scores (± standard deviations) obtained over 112 datasets with DFA, DRTP, and NoProp models. Results are averaged over the original split and five random seeds to ensure robustness and comparability.}
\begin{longtable*}{lccccc}
\toprule
\textbf{Name}                  & \textbf{Train / Test} & \textbf{Length / Class} & \textbf{DFA} & \textbf{DRTP} & \textbf{NoProp} \\ 
\midrule
\endfirsthead
\toprule
\textbf{Name}                  & \textbf{Train / Test} & \textbf{Length / Class} & \textbf{DFA} & \textbf{DRTP} & \textbf{NoProp} \\ 
\midrule
\endhead
\midrule
\multicolumn{6}{r}{Continued on next page} \\
\midrule
\endfoot
\bottomrule
\endlastfoot
ACSF1                                    & 100/100        & 1460/10    & 0.741 ± 0.036 & \textbf{0.788 ± 0.025} & 0.672 ± 0.052 \\
Adiac                                    & 390/391        & 176/37     & \textbf{0.754 ± 0.017} & 0.749 ± 0.018 & 0.708 ± 0.018 \\
ArrowHead                                & 36/175         & 251/3      & 0.810 ± 0.039 & \textbf{0.840 ± 0.050} & 0.828 ± 0.048 \\
Beef                                     & 30/30          & 470/5      & \textbf{0.760 ± 0.055} & 0.655 ± 0.085 & 0.649 ± 0.083 \\
BeetleFly                                & 20/20          & 512/2      & 0.790 ± 0.107 & 0.714 ± 0.129 & \textbf{0.802 ± 0.065} \\
BirdChicken                              & 20/20          & 512/2      & 0.668 ± 0.160 & 0.714 ± 0.131 & \textbf{0.789 ± 0.117} \\
BME                                      & 30/150         & 128/3      & 0.982 ± 0.010 & \textbf{0.992 ± 0.005} & 0.945 ± 0.039 \\
Car                                      & 60/60          & 577/4      & 0.812 ± 0.045 & \textbf{0.879 ± 0.021} & 0.828 ± 0.051 \\
CBF                                      & 30/900         & 128/3      & 0.905 ± 0.054 & 0.973 ± 0.025 & \textbf{0.979 ± 0.013} \\
Chinatown                                & 20/343         & 24/2       & 0.947 ± 0.033 & \textbf{0.968 ± 0.015} & 0.958 ± 0.013 \\
ChlorineConcentration                    & 467/3840       & 166/3      & \textbf{0.790 ± 0.019} & 0.690 ± 0.020 & 0.765 ± 0.015 \\
CinCECGTorso                             & 40/1380        & 1639/4     & 0.676 ± 0.065 & 0.869 ± 0.059 & \textbf{0.978 ± 0.012} \\
Coffee                                   & 28/28          & 286/2      & \textbf{0.982 ± 0.020} & 0.970 ± 0.027 & 0.928 ± 0.033 \\
Computers                                & 250/250        & 720/2      & 0.582 ± 0.026 & 0.645 ± 0.033 & \textbf{0.708 ± 0.057} \\
CricketX                                 & 390/390        & 300/12     & 0.685 ± 0.019 & 0.747 ± 0.018 & \textbf{0.804 ± 0.016} \\
CricketY                                 & 390/390        & 300/12     & 0.688 ± 0.023 & 0.767 ± 0.020 & \textbf{0.775 ± 0.034} \\
CricketZ                                 & 390/390        & 300/12     & 0.678 ± 0.015 & 0.756 ± 0.018 & \textbf{0.795 ± 0.012} \\
Crop                                     & 7200/16800     & 46/24      & \textbf{0.753 ± 0.006} & 0.743 ± 0.001 & 0.729 ± 0.004 \\
DiatomSizeReduction                      & 16/306         & 345/4      & 0.940 ± 0.027 & \textbf{0.943 ± 0.018} & 0.942 ± 0.027 \\
DistalPhalanxOutlineAgeGroup             & 400/139        & 80/3       & 0.748 ± 0.035 & \textbf{0.756 ± 0.017} & 0.725 ± 0.067 \\
DistalPhalanxOutlineCorrect              & 600/276        & 80/2       & \textbf{0.800 ± 0.029} & 0.797 ± 0.036 & 0.773 ± 0.027 \\
DistalPhalanxTW                          & 400/139        & 80/6       & 0.501 ± 0.042 & \textbf{0.529 ± 0.035} & 0.479 ± 0.049 \\
Earthquakes                              & 322/139        & 512/2      & \textbf{0.549 ± 0.032} & 0.501 ± 0.048 & 0.469 ± 0.036 \\
ECG200                                   & 100/100        & 96/2       & 0.864 ± 0.027 & \textbf{0.870 ± 0.042} & 0.834 ± 0.051 \\
ECG5000                                  & 500/4500       & 140/5      & 0.570 ± 0.016 & 0.564 ± 0.020 & \textbf{0.593 ± 0.021} \\
ECGFiveDays                              & 23/861         & 136/2      & \textbf{0.912 ± 0.047} & 0.896 ± 0.054 & 0.816 ± 0.076 \\
ElectricDevices                          & 8926/7711      & 96/7       & 0.760 ± 0.086 & 0.726 ± 0.082 & \textbf{0.803 ± 0.098} \\
EOGHorizontalSignal                      & 362/362        & 1250/12    & 0.595 ± 0.057 & 0.654 ± 0.076 & \textbf{0.693 ± 0.079} \\
EOGVerticalSignal                        & 362/362        & 1250/12    & 0.508 ± 0.097 & 0.579 ± 0.106 & \textbf{0.680 ± 0.081} \\
EthanolLevel                             & 504/500        & 1751/4     & 0.717 ± 0.056 & \textbf{0.763 ± 0.057} & 0.720 ± 0.060 \\
FaceAll                                  & 560/1690       & 131/14     & 0.922 ± 0.059 & \textbf{0.946 ± 0.065} & 0.939 ± 0.072 \\
FaceFour                                 & 24/88          & 350/4      & 0.909 ± 0.011 & \textbf{0.941 ± 0.020} & 0.932 ± 0.054 \\
FacesUCR                                 & 200/2050       & 131/14     & 0.875 ± 0.008 & \textbf{0.926 ± 0.014} & 0.894 ± 0.014 \\
FiftyWords                               & 450/455        & 270/50     & 0.561 ± 0.030 & \textbf{0.665 ± 0.023} & 0.347 ± 0.023 \\
Fish                                     & 175/175        & 463/7      & 0.865 ± 0.020 & \textbf{0.926 ± 0.023} & 0.892 ± 0.013 \\
FordA                                    & 3601/1320      & 500/2      & 0.886 ± 0.006 & 0.827 ± 0.053 & \textbf{0.956 ± 0.007} \\
FordB                                    & 3636/810       & 500/2      & 0.819 ± 0.061 & 0.784 ± 0.057 & \textbf{0.928 ± 0.038} \\
FreezerRegularTrain                      & 150/2850       & 301/2      & 0.989 ± 0.006 & \textbf{0.994 ± 0.006} & \textbf{0.994 ± 0.004} \\
FreezerSmallTrain                        & 28/2850        & 301/2      & 0.876 ± 0.048 & \textbf{0.890 ± 0.076} & 0.877 ± 0.048 \\
GunPoint                                 & 50/150         & 150/2      & 0.913 ± 0.042 & 0.953 ± 0.032 & \textbf{0.956 ± 0.029} \\
GunPointAgeSpan                          & 135/316        & 150/2      & 0.949 ± 0.011 & \textbf{0.961 ± 0.016} & 0.949 ± 0.018 \\
GunPointMaleVersusFemale                 & 135/316        & 150/2      & 0.980 ± 0.034 & \textbf{0.997 ± 0.003} & \textbf{0.997 ± 0.005} \\
GunPointOldVersusYoung                   & 136/315        & 150/2      & 0.997 ± 0.008 & \textbf{1.000 ± 0.000} & \textbf{1.000 ± 0.000} \\
Ham                                      & 109/105        & 431/2      & \textbf{0.830 ± 0.044} & 0.800 ± 0.047 & 0.748 ± 0.039 \\
HandOutlines                             & 1000/370       & 2709/2     & 0.900 ± 0.014 & \textbf{0.912 ± 0.011} & 0.886 ± 0.013 \\
Haptics                                  & 155/308        & 1092/5     & 0.426 ± 0.025 & \textbf{0.445 ± 0.012} & 0.429 ± 0.025 \\
Herring                                  & 64/64          & 512/2      & 0.586 ± 0.022 & 0.518 ± 0.032 & \textbf{0.613 ± 0.058} \\
HouseTwenty                              & 40/119         & 2000/2     & 0.755 ± 0.042 & \textbf{0.879 ± 0.016} & 0.867 ± 0.084 \\
InlineSkate                              & 100/550        & 1882/7     & 0.334 ± 0.015 & \textbf{0.340 ± 0.018} & 0.339 ± 0.014 \\
InsectEPGRegularTrain                    & 62/249         & 601/3      & \textbf{1.000 ± 0.000} & \textbf{1.000 ± 0.000} & 0.995 ± 0.013 \\
InsectEPGSmallTrain                      & 17/249         & 601/3      & \textbf{1.000 ± 0.000} & \textbf{1.000 ± 0.000} & 0.991 ± 0.022 \\
InsectWingbeatSound                      & 220/1980       & 256/11     & 0.604 ± 0.025 & \textbf{0.641 ± 0.020} & 0.620 ± 0.019 \\
ItalyPowerDemand                         & 67/1029        & 24/2       & 0.953 ± 0.007 & \textbf{0.964 ± 0.004} & 0.953 ± 0.013 \\
LargeKitchenAppliances                   & 375/375        & 720/3      & 0.619 ± 0.020 & 0.680 ± 0.038 & \textbf{0.886 ± 0.033} \\
Lightning2                               & 60/61          & 637/2      & 0.716 ± 0.044 & \textbf{0.758 ± 0.043} & 0.749 ± 0.029 \\
Lightning7                               & 70/73          & 319/7      & 0.658 ± 0.059 & \textbf{0.701 ± 0.031} & 0.681 ± 0.034 \\
Mallat                                   & 55/2345        & 1024/8     & 0.936 ± 0.024 & \textbf{0.969 ± 0.027} & 0.930 ± 0.037 \\
Meat                                     & 60/60          & 448/3      & \textbf{0.986 ± 0.026} & \textbf{0.986 ± 0.016} & 0.967 ± 0.034 \\
MedicalImages                            & 381/760        & 99/10      & 0.693 ± 0.014 & 0.697 ± 0.022 & \textbf{0.703 ± 0.025} \\
MiddlePhalanxOutlineAgeGroup             & 400/154        & 80/3       & 0.604 ± 0.085 & \textbf{0.622 ± 0.103} & 0.611 ± 0.105 \\
MiddlePhalanxOutlineCorrect              & 600/291        & 80/2       & 0.828 ± 0.009 & 0.809 ± 0.020 & \textbf{0.832 ± 0.035} \\
MiddlePhalanxTW                          & 399/154        & 80/6       & 0.397 ± 0.025 & 0.368 ± 0.027 & \textbf{0.407 ± 0.040} \\
MixedShapesRegularTrain                  & 500/2425       & 1024/5     & 0.904 ± 0.009 & 0.928 ± 0.007 & \textbf{0.931 ± 0.002} \\
MixedShapesSmallTrain                    & 100/2425       & 1024/5     & 0.809 ± 0.012 & 0.839 ± 0.008 & \textbf{0.850 ± 0.015} \\
MoteStrain                               & 20/1252        & 84/2       & \textbf{0.864 ± 0.019} & 0.860 ± 0.034 & 0.825 ± 0.030 \\
NonInvasiveFetalECGThorax1               & 1800/1965      & 750/42     & \textbf{0.927 ± 0.005} & 0.919 ± 0.004 & 0.905 ± 0.010 \\
NonInvasiveFetalECGThorax2               & 1800/1965      & 750/42     & \textbf{0.936 ± 0.005} & 0.933 ± 0.004 & 0.923 ± 0.008 \\
OliveOil                                 & 30/30          & 570/4      & 0.871 ± 0.085 & \textbf{0.881 ± 0.086} & 0.797 ± 0.092 \\
OSULeaf                                  & 200/242        & 427/6      & 0.528 ± 0.022 & 0.705 ± 0.035 & \textbf{0.708 ± 0.033} \\
PhalangesOutlinesCorrect                 & 1800/858       & 80/2       & \textbf{0.832 ± 0.009} & 0.801 ± 0.017 & 0.825 ± 0.016 \\
Phoneme                                  & 214/1896       & 1024/39    & 0.086 ± 0.002 & 0.119 ± 0.007 & \textbf{0.133 ± 0.015} \\
PigAirwayPressure                        & 104/208        & 2000/52    & \textbf{0.093 ± 0.019} & 0.087 ± 0.014 & 0.072 ± 0.012 \\
PigArtPressure                           & 104/208        & 2000/52    & 0.190 ± 0.022 & 0.182 ± 0.026 & \textbf{0.229 ± 0.032} \\
PigCVP                                   & 104/208        & 2000/52    & 0.109 ± 0.022 & \textbf{0.114 ± 0.021} & 0.111 ± 0.014 \\
Plane                                    & 105/105        & 144/7      & 0.983 ± 0.007 & \textbf{0.997 ± 0.005} & 0.990 ± 0.009 \\
PowerCons                                & 180/180        & 144/2      & 0.994 ± 0.009 & \textbf{0.995 ± 0.006} & 0.988 ± 0.014 \\
ProximalPhalanxOutlineAgeGroup           & 400/205        & 80/3       & 0.788 ± 0.029 & \textbf{0.799 ± 0.024} & 0.778 ± 0.029 \\
ProximalPhalanxOutlineCorrect            & 600/291        & 80/2       & \textbf{0.870 ± 0.014} & 0.859 ± 0.011 & 0.860 ± 0.009 \\
ProximalPhalanxTW                        & 400/205        & 80/6       & \textbf{0.594 ± 0.042} & 0.535 ± 0.040 & 0.571 ± 0.025 \\
RefrigerationDevices                     & 375/375        & 720/3      & 0.482 ± 0.022 & 0.571 ± 0.041 & \textbf{0.633 ± 0.081} \\
Rock                                     & 20/50          & 2844/4     & 0.811 ± 0.096 & \textbf{0.815 ± 0.060} & 0.752 ± 0.113 \\
ScreenType                               & 375/375        & 720/3      & 0.468 ± 0.028 & 0.502 ± 0.033 & \textbf{0.508 ± 0.052} \\
SemgHandGenderCh2                        & 300/600        & 1500/2     & 0.927 ± 0.023 & 0.855 ± 0.034 & \textbf{0.935 ± 0.019} \\
SemgHandMovementCh2                      & 450/450        & 1500/6     & \textbf{0.791 ± 0.032} & 0.764 ± 0.032 & 0.736 ± 0.031 \\
SemgHandSubjectCh2                       & 450/450        & 1500/5     & \textbf{0.943 ± 0.017} & 0.885 ± 0.017 & 0.835 ± 0.042 \\
ShapeletSim                              & 20/180         & 500/2      & 0.533 ± 0.031 & 0.572 ± 0.033 & \textbf{0.637 ± 0.188} \\
ShapesAll                                & 600/600        & 512/60     & 0.743 ± 0.008 & \textbf{0.790 ± 0.008} & 0.786 ± 0.012 \\
SmallKitchenAppliances                   & 375/375        & 720/3      & 0.708 ± 0.024 & \textbf{0.764 ± 0.012} & 0.738 ± 0.007 \\
SmoothSubspace                           & 150/150        & 15/3       & 0.960 ± 0.019 & \textbf{0.972 ± 0.023} & 0.950 ± 0.018 \\
SonyAIBORobotSurface1                    & 20/601         & 70/2       & 0.866 ± 0.079 & \textbf{0.938 ± 0.058} & 0.912 ± 0.039 \\
SonyAIBORobotSurface2                    & 27/953         & 65/2       & \textbf{0.877 ± 0.025} & 0.852 ± 0.051 & 0.860 ± 0.053 \\
StarLightCurves                          & 1000/8236      & 1024/3     & 0.928 ± 0.005 & 0.957 ± 0.005 & \textbf{0.960 ± 0.006} \\
Strawberry                               & 613/370        & 235/2      & \textbf{0.972 ± 0.007} & 0.962 ± 0.011 & 0.968 ± 0.009 \\
SwedishLeaf                              & 500/625        & 128/15     & 0.918 ± 0.008 & 0.935 ± 0.010 & \textbf{0.948 ± 0.007} \\
Symbols                                  & 25/995         & 398/6      & 0.844 ± 0.015 & 0.893 ± 0.015 & \textbf{0.899 ± 0.018} \\
SyntheticControl                         & 300/300        & 60/6       & 0.976 ± 0.005 & \textbf{0.996 ± 0.004} & 0.987 ± 0.007 \\
ToeSegmentation1                         & 40/228         & 277/2      & 0.590 ± 0.040 & 0.889 ± 0.024 & \textbf{0.920 ± 0.037} \\
ToeSegmentation2                         & 36/130         & 343/2      & 0.614 ± 0.041 & 0.733 ± 0.068 & \textbf{0.880 ± 0.058} \\
Trace                                    & 100/100        & 275/4      & 0.949 ± 0.034 & 0.988 ± 0.015 & \textbf{1.000 ± 0.000} \\
TwoLeadECG                               & 23/1139        & 82/2       & 0.906 ± 0.045 & \textbf{0.930 ± 0.047} & 0.901 ± 0.095 \\
TwoPatterns                              & 1000/4000      & 128/4      & 0.922 ± 0.006 & 0.998 ± 0.001 & \textbf{1.000 ± 0.000} \\
UMD                                      & 36/144         & 150/3      & \textbf{0.939 ± 0.026} & 0.853 ± 0.025 & 0.900 ± 0.023 \\
UWaveGestureLibraryAll                   & 896/3582       & 945/8      & 0.948 ± 0.002 & 0.964 ± 0.002 & \textbf{0.975 ± 0.002} \\
UWaveGestureLibraryX                     & 896/3582       & 315/8      & 0.761 ± 0.012 & 0.808 ± 0.008 & \textbf{0.826 ± 0.006} \\
UWaveGestureLibraryY                     & 896/3582       & 315/8      & 0.700 ± 0.008 & 0.719 ± 0.009 & \textbf{0.745 ± 0.011} \\
UWaveGestureLibraryZ                     & 896/3582       & 315/8      & 0.715 ± 0.006 & 0.733 ± 0.008 & \textbf{0.757 ± 0.007} \\
Wafer                                    & 1000/6164      & 152/2      & 0.990 ± 0.003 & 0.988 ± 0.004 & \textbf{0.992 ± 0.002} \\
Wine                                     & 57/54          & 234/2      & \textbf{0.892 ± 0.090} & 0.848 ± 0.115 & 0.829 ± 0.100 \\
WordSynonyms                             & 267/638        & 270/25     & 0.477 ± 0.017 & 0.536 ± 0.017 & \textbf{0.573 ± 0.029} \\
Worms                                    & 181/77         & 900/5      & 0.496 ± 0.037 & 0.585 ± 0.061 & \textbf{0.665 ± 0.050} \\
WormsTwoClass                            & 181/77         & 900/2      & 0.596 ± 0.046 & 0.643 ± 0.023 & \textbf{0.737 ± 0.047} \\
Yoga                                     & 300/3000       & 426/2      & 0.850 ± 0.015 & 0.839 ± 0.018 & \textbf{0.861 ± 0.010} \\
\bottomrule
\end{longtable*}}

\clearpage
\twocolumn

\end{document}